\documentclass[11pt]{article}

\usepackage[final]{acl}

\usepackage{times}
\usepackage{latexsym}
\usepackage{hyperref}
\usepackage{url}
\usepackage{booktabs}
\usepackage{makecell}
\usepackage{graphicx} 
\usepackage{multirow}
\usepackage{graphicx}
\usepackage{subcaption}
\usepackage{wrapfig} 
\usepackage{booktabs} 
\usepackage[table]{xcolor} 
\usepackage{wrapfig}
\usepackage{xurl}
\usepackage{amsmath}
 \usepackage{algorithm,algpseudocode}
\usepackage{amssymb} 
\usepackage{bbding}    
\usepackage{pifont}     
\usepackage{wasysym}    
\usepackage{booktabs}   
\usepackage{tabularx}   
\usepackage{amsmath}    
\definecolor{textblue}{RGB}{70,130,180}
\definecolor{textred}{RGB}{205,92,92}
\definecolor{lightblue}{rgb}{0.8, 0.9, 1}
\definecolor{lightpink}{RGB}{255,192,203}
\definecolor{lightblue}{rgb}{0.8, 0.9, 1}
\definecolor{lightyellow}{RGB}{255,255,224}
\definecolor{lightgreen}{RGB}{144,238,144}
\definecolor{gray}{RGB}{220 220 220}
\usepackage{enumitem}
\usepackage{adjustbox}

\usepackage[T1]{fontenc}

\usepackage[utf8]{inputenc}

\usepackage{microtype}

\usepackage{inconsolata}

\usepackage{graphicx}

\title{FACT-E: Causality-Inspired Evaluation for Trustworthy \\ Chain-of-Thought Reasoning}

\author{
  Yuxi Sun\textsuperscript{1} \quad 
  Aoqi Zuo\textsuperscript{2} \quad 
  Haotian Xie\textsuperscript{1} \quad 
  Wei Gao\textsuperscript{3} \quad 
  Mingming Gong\textsuperscript{2} \quad 
  Jing Ma\textsuperscript{1}\thanks{Corresponding author.} \\
  \textsuperscript{1}Hong Kong Baptist University 
  \textsuperscript{2}The University of Melbourne 
  \textsuperscript{3}Singapore Management University \\
  {\texttt{\{csyxsun, jingma\}@comp.hkbu.edu.hk}}\quad { \texttt{azuo@student.unimelb.edu.au}}\\
   { \texttt{weigao@smu.edu.sg}}\quad { \texttt{mingming.gong@unimelb.edu.au}} 
}

\begin{document}
\maketitle
\begin{abstract}
Chain-of-Thought (CoT) prompting has improved LLM reasoning, but models often generate explanations that appear coherent while containing unfaithful intermediate steps. Existing self-evaluation approaches are prone to inherent biases: the model may confidently endorse coherence even when the step-to-step implication is not valid, leading to unreliable faithfulness evaluation. We propose FACT-E, a causality-inspired framework for evaluating CoT quality. FACT-E uses controlled perturbations as an instrumental signal to separate genuine step-to-step dependence from bias-driven artifacts, producing more reliable faithfulness estimates (\textit{intra-chain faithfulness}). To select trustworthy trajectories, FACT-E jointly considers \textit{intra-chain faithfulness} and \textit{CoT-to-answer consistency}, ensuring that selected chains are both faithful internally and supportive of the correct final answer. Experiments on GSM8K, MATH, and CommonsenseQA show that FACT-E improves reasoning-trajectory selection and yields stronger in-context learning exemplars. FACT-E also reliably detects flawed reasoning under noisy conditions, providing a robust metric for trustworthy LLM reasoning.

\end{abstract}

\section{Introduction}
The paradigm of Chain-of-Thought (CoT) prompting has fundamentally enhanced the reasoning capabilities of Large Language Models (LLMs)~\citep{wei2022chain,yu2025causalsufficiencynecessityimproves,fu2025unveiling}. However, a critical challenge persists in discerning the reliability of reasoning trajectories~\citep{sun2025causalabstain}. Models frequently generate rationales that yield correct results and appear superficially persuasive~\citep{cui2024theoretical, turpin2024language, lanham2023measuring}, yet are fundamentally intra-chain unfaithful, characterized by \textit{broken logical dependencies between intermediate steps or the inclusion of inaccurate content}. Detecting such intra-chain unfaithfulness is thus crucial for improving the robustness and trustworthiness of model-generated reasoning.

\begin{figure}[t]
\centering
\includegraphics[width=1\linewidth]{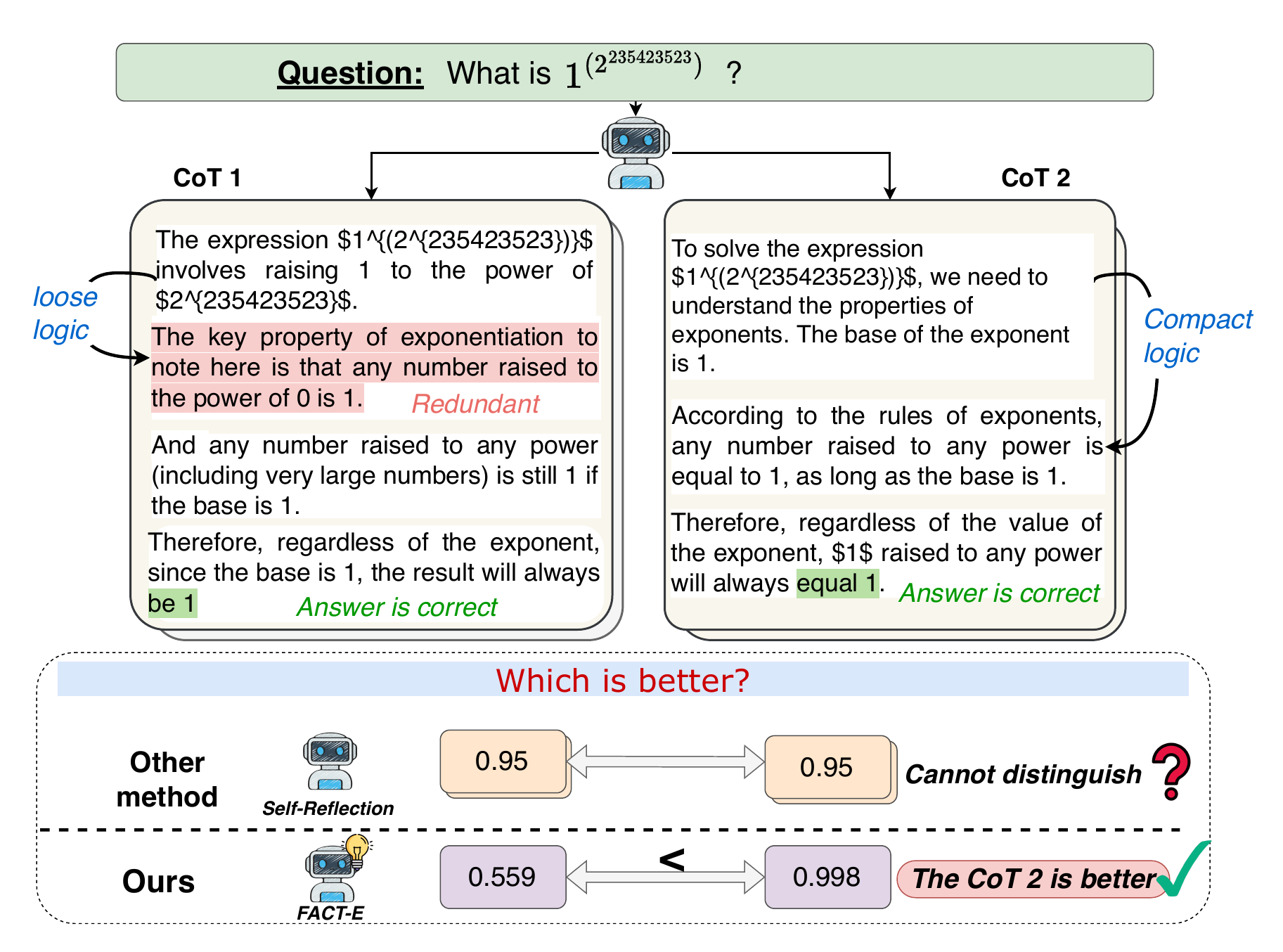}
\caption{Motivating example illustrating the limitation of LLM self-assessment on CoT evaluation.
Two reasoning chains appear fluent and coherent, yet CoT~1 contains successive intermediate steps that are not logically necessary for subsequent reasoning. Conventional method (e.g., self-reflect)
assigns similarly high quality scores to both chains, failing to detect this breakdown, whereas FACT-E evaluates the unfaithfulness in a chain and successfully identifies CoT~2 as more trustworthy.}
\vspace{-1.2em}
\label{fig:exp1}
\end{figure}

Existing methodologies for recognizing faithful and filtering trustworthy reasoning traces generally fall into two paradigms. (1) \textit{LLM-as-Judge Methods} leverage the model itself as an evaluator. Techniques such as self-correction and self-reflection operate in a black-box manner to assess whether a generated CoT supports the final answer~\citep{kadavath2022languagemodelsmostlyknow,xi2024selfpolishenhancereasoninglarge,madaan2023self}. Another line of work decomposes CoT into sub-questions and verifies intermediate steps against corresponding sub-answers~\cite{radhakrishnan2023question,zhu-etal-2023-chain}. 
Crucially, this paradigm is inherently \textit{answer-centric}: it relies primarily on changes in the final answer as supervision, overlooking the internal dependencies among intermediate reasoning steps, which implicitly assume that such surface-level correction implies logical validity.
(2) \textit{Causal-based Methods} have subsequently emerged, employing causal interventions to evaluate reasoning quality. Some approaches assess whether a reasoning chain is faithful to the final answer by perturbing inputs or constructing counterfactuals~\cite{yang2025well,xiong2025measuring}. Others leverage causal measures such as the Probability of Necessity and Sufficiency (PNS) to identify redundant or non-influential steps, or task LLMs with autonomously generating causal graphs to support structured reasoning~\citep{yu2025causalsufficiencynecessityimproves,hüyük2025reasoningelicitationlanguagemodels,fu2025unveiling}.

Despite their causal underpinnings, the efficacy of these methods is hindered by the inherent biases of LLM-based evaluators~\citep{fu2025unveiling,yu2025causalsufficiencynecessityimproves}. This dependence induces a closed-loop feedback fallacy~\citep{huang2023large,zheng2023judging}, wherein an LLM may confidently validate the faithfulness between its generated reasoning steps despite the absence of a rigorous logical entailment~\citep{jiang2024peek,mckenna2023sources,zheng2023judging,huang2023large}. Take the question in Figure~\ref{fig:exp1} as an example, 
in CoT 1, the second step (``\textit{The key property of exponentiation to note here is that any number raised to the power of 0 is 1}.'') logically deviates from the first step in the reasoning path (``\textit{The expression $1^{2^{235423523}}$ involves raising 1 to the power of $2^{235423523}$}''). However, traditional LLM evaluation methods (e.g., self-reflection~\citep{kadavath2022language}) may struggle to distinguish logically sparse reasoning between these two successive steps in a CoT. As a result, they assign similar quality scores to CoT 1 and CoT 2, 
despite the substantial difference in their internal coherence. This issue may stem from spurious correlation between the LLM's assessment and its internal bias (e.g., LLMs demonstrate a persistent self-affirmation bias, consistently assigning positive evaluations to their own generations with negligible variance~\citep{huang2023large}). Consequently, such spurious correlations can make the model overconfident or cause it to neglect evaluating the faithfulness between segments, regardless of their true relationship.

In this work, we propose a novel causal view based on a Structural Causal Model (SCM)~\citep{pearl2009causality} to support the LLM self-assessment of intra-chain faithfulness via a contrastive design. To mitigate spurious correlations between the LLM’s faithfulness assessment and its internal biases, we introduce external noise as an instrumental variable, yielding a more reliable faithfulness evaluation. In addition to intra-chain faithfulness, we also consider answer correctness as a complementary dimension of CoT quality. Accordingly, we introduce FACT-E (\underline{F}aithfulness \underline{A}nd \underline{C}onsistency \underline{T}andem \underline{E}stimation), a causality-inspired framework for CoT quality estimation. FACT-E consists of two modules: \textit{CoT-to-Answer Consistency}, which verifies that the reasoning chain supports the correct final answer, and \textit{Intra-Chain Faithfulness}, which leverages causal insights to refine the LLM’s faithfulness judgments.

We empirically validate FACT-E efficacy across mathematical and commonsense reasoning tasks. Our results show that selecting reasoning paths based on FACT-E scores substantially improves answer accuracy and enhances in-context learning. Furthermore, experiments under noisy conditions demonstrate that our approach effectively identifies process-level failures, thereby improving the robustness and controllability of LLM reasoning.

Our contributions are mainly threefold:
\begin{itemize}[leftmargin=*]
\item We leverage causality to obtain more reliable intra-chain faithfulness evaluations by introducing external noise as an instrumental variable, mitigating the impact of unobserved LLM biases in self-assessment.
\item We propose FACT-E, a novel CoT evaluation framework that jointly considers (i) answer correctness implied by the CoT (\textit{CoT-to-Answer Consistency}) and (ii) faithfulness between successive CoT segments (\textit{Intra-Chain Faithfulness}).
\item We conduct experiments across three representative tasks: (1) \textit{Improving Answer Accuracy} by selecting higher-quality CoT; (2) \textit{Enhancing In-Context Learning} by using optimized chains as exemplars; and (3) \textit{Noise Detection} by identifying flawed reasoning. The results show that FACT-E achieves competitive performance against strong baselines.
\end{itemize}

\section{Task Formulation with A Causal View}
In this section, we formally formulate the task of measuring the trustworthiness of a reasoning chain from a causal perspective.
\begin{figure}[t!]
    \centering
    \includegraphics[width=0.95\linewidth]{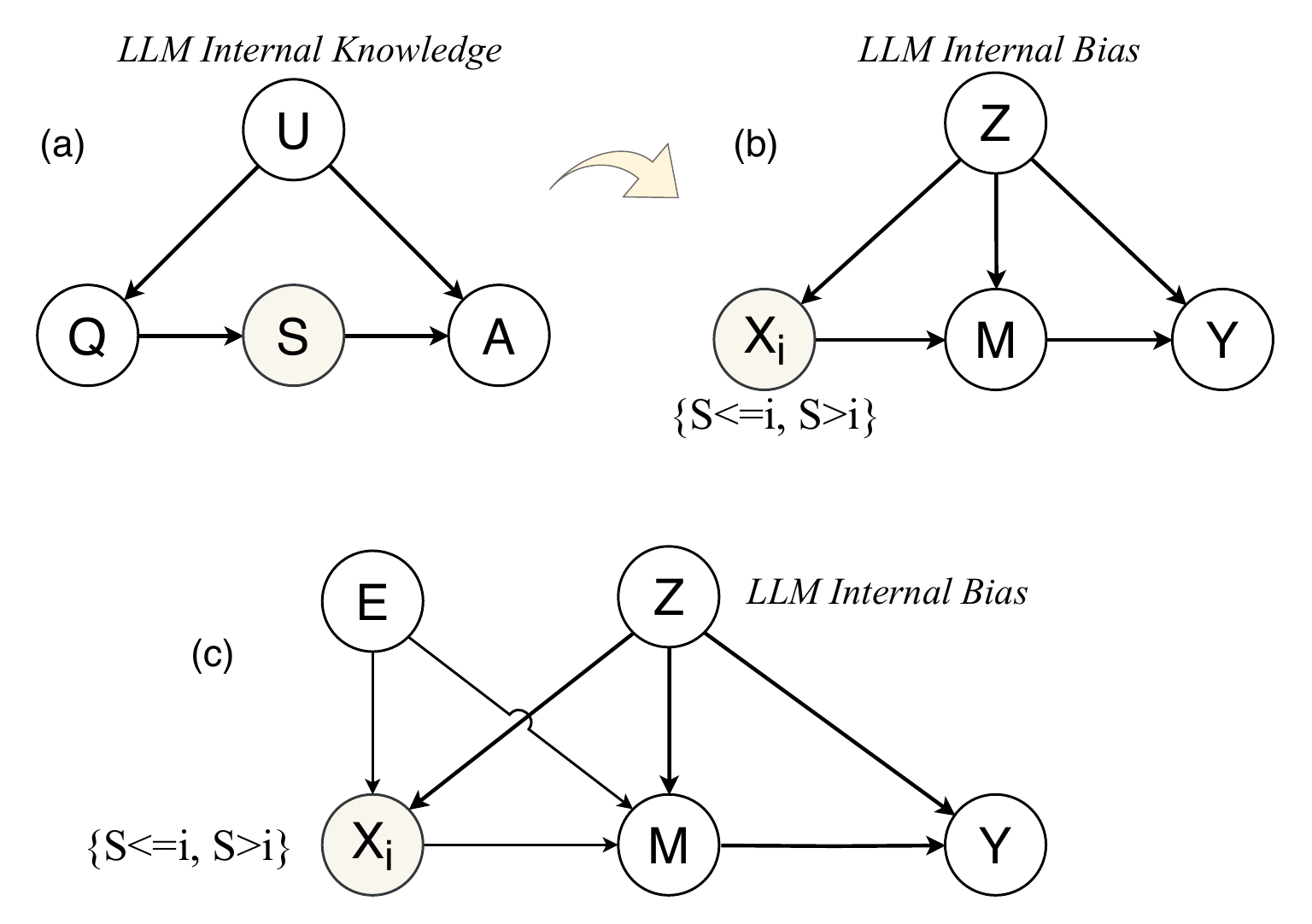}
    \caption{Structural causal graphs for CoT quality estimation. (a) depicts the process of answering a query $Q$ with CoT $\mathbf{S}$. (b) illustrates the traditional self-assessment approaches, where the LLM evaluates the faithfulness score $Y$ between \( \mathbf{S}_{\leq i} \) and  \( \mathbf{S}_{> i} \), denoted by $X_i$, mediated by LLM's self-evaluation \( M \). The LLM's internal bias \( Z \) is an unobserved confounder that affects all variables. (c) FACT-E introduces exogenous noise $E$ as an instrumental variable to obtain a more reliable faithfulness evaluation for CoTs.}
    \vspace{-1em}
\label{graph}
\end{figure}

\subsection{Problem Definition}
Given a query $Q$, the LLM is prompted to generate a set of reasoning chains $\mathcal{S}$. To effectively filter and select trustworthy reasoning chains from a pool of post-hoc candidates, our goal is to estimate a reliability score $\mathcal{R}_{\mathbf{S}}$ for each candidate chain $\mathbf{S} \in \mathcal{S}$ as a quality measure, denoted as ${\text{LLM}}(Q, \mathbf{S}) \to \mathcal{R}_{\mathbf{S}} \in [0, 1]$. A higher $\mathcal{R}_{\mathbf{S}}$ indicates that the reasoning process is not only correct in its final outcome but also faithful among its intermediate steps. We model these two aspects below.
\subsection{CoT-to-Answer Consistency}

A correct answer is a prerequisite for a high-quality CoT. Accordingly, we first model the chain’s consistency with the correct outcome.
\paragraph{Definition 1 (CoT-to-Answer Consistency).} \textit{An evaluative metric that quantifies the trustworthiness of a CoT candidate by estimating the probability that the reasoning chain consistently leads to the correct final answer.}

Following recent work~\citep{wu-etal-2024-decot,fu2025unveiling,zhang2024causal}, we model the reasoning process using Structural Causal Model (SCM) as $Q \rightarrow \mathbf{S} \rightarrow A$, where 
the CoT $\mathbf{S}$ acts as a mediator that transmits the influence of the query $Q$ to the final answer $A$. Besides, since the reasoning process can be causally dependent on LLM's internal knowledge $U$, we model this process in Figure~\ref{graph}(a). A robust $\mathcal{R}_{\mathbf{S}}$ must capture whether the mediation is consistently reaching the correct $A$.

\subsection{Intra-Chain Faithfulness}
Merely ensuring that a CoT yields the correct answer is insufficient for a comprehensive reliability assessment. A correct answer can often be reached through flawed or hallucinated reasoning steps. 
To capture the reliability of the intermediate steps beyond the final answer, we introduce Intra-Chain Faithfulness.
\paragraph{Definition 2 (Intra-Chain Faithfulness).} \textit{A measurement of a reasoning path’s quality, focusing on the logical dependencies between steps and their content correctness. A failure of faithfulness occurs when a CoT appears superficially coherent but relies on fragile connections or exhibits erroneous intermediate steps.}

Formally, let $\mathbf{S} = \{s_1, s_2, \ldots, s_L\}$ denotes a CoT consisting of $L$ steps. 
To evaluate the faithfulness degree in the chain, we decompose the reasoning process at step $i$ into successive two segments, $\mathbf{S}_{\leq i}$ and $\mathbf{S}_{> i}$, denoted by $X_i$, where the segment $\mathbf{S}_{\leq i} = \{s_1, \ldots, s_i\}$ precedes the subsequent segment $\mathbf{S}_{> i} = \{s_{i+1}, \ldots, s_L\}$. In Figure~\ref{graph}(b), we denote the faithfulness score between $\mathbf{S}_{\leq i}$ and $\mathbf{S}_{> i}$ by $Y$. The mediator $M$ represents the LLM's evaluation on their faithfulness. Ideally, a robust faithfulness estimation is achieved when the mediation process $M$ remains unbiased. However, in practice, LLM evaluation is affected by its unobservable internal bias $Z$ (e.g., self-affirmation bias). The overall faithfulness score of a CoT is aggregated as the faithfulness score across all split indices $i$ from $1$ to $L-1$.

\section{Methodology}\label{method}
\subsection{Quantify Intra-Chain Faithfulness}
Ideally, LLM assessment should act as an objective judge of faithfulness. However, in practice, the internal bias may exist. Specifically, two major sources of bias arise: 
(1) \textit{Self-affirmation bias}, where the LLM exhibits an inherent tendency to positively evaluate its own outputs~\citep{huang2023large,zheng2023judging}; and (2) \textit{Statistical shortcuts}, where the LLM relies on shallow heuristics learned during pre-training, such as lexical overlap or frequent co-occurrence patterns. As a result, the LLM may hallucinate strong logical connections based on statistical familiarity, even when the underlying reasoning progression is flawed~\citep{zheng2023judging,xiong2025measuring,jiang-etal-2024-peek}. 

As illustrated in Figure~\ref{graph}(b), there exists a spurious correlation between LLM assessment on the faithfulness degree between intermediate steps in a CoT due to the unobservable LLM's internal bias.

To address this issue, we introduce an exogenous instrumental variable $E$. Since we cannot directly intervene on the CoT generation process, we approximate causal interventions by using $E$ to construct counterfactual segments and inject them into the chain. Concretely, $E$ denotes randomly injected perturbations that disrupt logical dependencies (e.g., omitting steps) and corrupt content correctness (e.g., introducing operation errors). Appendix~\ref{noise_inject} lists a non-exhaustive set of noise configurations. Since these perturbations modify $\mathbf{S}_{> i}$ and thus the faithfulness relation between $\mathbf{S}_{\leq i}$ and $\mathbf{S}_{> i}$, $E$ causally influences both $X_i$ and $M$. Figure~\ref{graph}(c) illustrates this intervention process.

By intervening on $E$, we observe corresponding changes in the assessed faithfulness $Y$ through the mediator $M$. The Average Causal Effect (ACE) directly characterizes faithfulness by measuring its sensitivity to structured perturbations. For each noise type $e_j \in \mathcal{E}$, given $X_i$ and its inference process $M$, we define ACE as:

\begin{equation}
    \begin{aligned}
    \nonumber
        \mathrm{ACE}(e_j, X_i) &= \mathbb{E}[Y \mid X_i, M, do(E=\emptyset)] \\&\quad - \mathbb{E}[Y \mid X_i, M, do(E=e_j)].
    \end{aligned}
\end{equation}
Intuitively, this quantity aligns with intuition: the ACE measures the relatively increased confidence of LLM evaluation $M$ on the faithfulness $Y$ of ${X}_{i}$ with respect to the ${X}_{i}$ injected with noises $e_j$.
We propose to measure the average causal effect across all types of intervened perturbations $\mathcal{E}$:
\[
\mathrm{ACE}(\mathcal{E}, X_i) = \frac{1}{|\mathcal{E}|}\sum_{e_j \in \mathcal{E}}\mathrm{ACE}(e_j, X_i).
\]
\paragraph{Estimation with contrastive design.} To estimate ACE, we introduce a contrastive design between the original and injected-noise reasonings. Specifically, we implement $\mathcal{E}$ as functional interventions on the LLM’s rollout process, instead of static noise injection. For each segmentation point $i$, we define a perturbed counterpart $X_i^{(e_j)} = (\mathbf{S}_{\leq i}, \mathbf{S}'_{> i})$, where $\mathbf{S}'_{> i}$ is a counterfactual rollout trajectory generated by LLM conditioned on both prefix $\mathbf{S}_{\leq i}$ and a specific logical perturbation $e_j \in \mathcal{E}$. Table~\ref{tab:exp} illustrates such a perturbed rollout, where a specific operation error is injected while maintaining the original prefix. The ACE thus quantifies the relative confidence increase on the original reasoning continuation $\mathbf{S}_{> i}$ compared to its perturbed counterpart $\mathbf{S}'_{> i}$, denoted by $\mathcal{F}_{\mathbf{S}}$:
\begin{equation}
\small
\nonumber
    \begin{aligned}
    \mathcal{F}_{\mathbf{S}} &= \frac{1}{|\mathcal{E}|(L-1)} \sum_{i=1}^{L-1} \sum_{e_j \in \mathcal{E}} \mathrm{ACE}(e_j, X_i) \\
    &\approx \frac{1}{|\mathcal{E}|(L-1)} \sum_{i=1}^{L-1} \sum_{e_j \in \mathcal{E}} 
    \mathbf{1}\!\left[(\mathbf{S}_{\leq i}, \mathbf{S}_{> i}) \succ (\mathbf{S}_{\leq i}, \mathbf{S}'^{(e_j)}_{> i})\right],
    \end{aligned}
\end{equation}
where $L$ denotes the number of steps in the CoT, and $\mathbf{1}[\cdot]$ is an indicator function that equals $1$ if the original $\mathbf{S}_{> i}$ is preferred over the perturbed $\mathbf{S}'_{> i}$ given the same prefix $\mathbf{S}_{\leq i}$, and $0$ otherwise.

This contrastive design serves two critical purposes. First, it effectively neutralizes \textit{self-affirmation bias}. Since both trajectories are self-generated by an LLM and share an identical prefix and stylistic context, the model cannot rely on stylistic familiarity or surface-level patterns—features that typically lead LLMs to assign inflated scores to their own outputs. Second, the injection of logical noise $e_j$ acts as a ``stress test.'' Even when the reasoning remains linguistically fluent and sound, its logical chain is intentionally fractured. This forces the LLM to look past \textit{superficial patterns} and focus on the \textit{actual reasoning steps}. Consequently, our method improves the intra-chain faithfulness score in reflecting true logical integrity rather than mere plausible-sounding text.

\begin{table}[t!]
\centering
\scriptsize
\setlength{\tabcolsep}{3pt}
\renewcommand{\arraystretch}{0.9}
\begin{tabularx}{\linewidth}{@{} l >{\raggedright\arraybackslash}X @{}}
\toprule
\textbf{Question} & 
\multicolumn{1}{>{\raggedright}p{0.88\linewidth}}{Suppose $\sin D = 0.7$ in the right triangle $DEF$ (where $\angle E = 90^\circ$ and $EF = 7$). What is the length of $DE$?} \\
\midrule
$(\mathbf{S}_{\leq i}, \mathbf{S}_{> i})$ &
We are given that $\sin D = 0.7$ and need to find the length of $DE$. In right triangle $DEF$, $EF=7$ is opposite to angle $D$, and $DF$ is the hypotenuse. \textcolor{purple}{Thus, $0.7 = \frac{7}{DF}$, so $DF=10$. By the Pythagorean theorem, $DE^2 = DF^2 - EF^2 = 100 - 49 = 51$, hence $DE=\sqrt{51}$.} \\

$(\mathbf{S}_{\leq i}, \mathbf{S}'_{> i})$ &
We are given that $\sin D = 0.7$ and need to find the length of $DE$. In right triangle $DEF$, $EF=7$ is opposite to angle $D$, and $DF$ is the hypotenuse. 
\textcolor{purple}{Thus, $\sin D = \frac{EF}{DE}$, so $DE=\frac{7}{0.7}=10$.} \\
\bottomrule
\end{tabularx}
\vspace{-1em}
\caption{Illustrative example of a perturbed rollout $\mathbf{S}'_{> i}$ generated from the same prefix $\mathbf{S}_{\leq i}$. The injected logical error leads to an incorrect continuation despite identical contextual and stylistic conditions.}
\label{tab:exp}
\end{table}

\subsection{Quantify CoT-to-Answer Consistency} 
Definition~1 on \textit{CoT-to-Answer Consistency} quantifies the model's confidence on a given reasoning chain $\mathbf{S}$ yielding the correct final answer. 
Formally, given a query $Q = q$ and a candidate reasoning chain $\mathbf{S}$ (we use $Q$ to denote the query random variable and $q$ to denote a concrete query instance), the LLM is prompted $ N$ times independently to judge whether $\mathbf{S}$ is sufficient to reach the correct answer. For each trial $n$, we define $J^{(n)} \in \{0, 1\}$ as the model's binary judgment, where $J^{(n)}=1$ indicates that the model judges $\mathbf{S}$ yields the correct answer, and $J^{(n)}=0$ otherwise. Let $P\!\left(J^{(n)}=1 \mid q,\mathbf{S}\right)$
denote the model-estimated probability of the positive judgment. We compute the CoT-to-Answer Consistency score by averaging these probabilities across the trials where the model explicitly validates the reasoning:
\begin{equation}
\nonumber
    \begin{aligned}
        \mathcal{C}_{\mathbf{S}} 
        \approx & \frac{1}{N} \sum_{n=1}^{N} P(J^{(n)} = 1 \mid q, \mathbf{S}) \cdot \mathbf{1}[J^{(n)} = 1],
    \end{aligned}
\end{equation}
where $\mathbf{1}[\cdot]$ is the indicator function. A high $\mathcal{C}_{\mathbf{S}}$ indicates consistent and confident prediction that the reasoning chain $\mathbf{S}$ provides a viable path to the correct answer, independent of explicit verification of intermediate logical validity.

\subsection{Tandem Estimation}\label{def:cot}
A high-quality CoT should not only arrive at the correct answer but also ensure that its intermediate steps are faithful. When faithfulness ($\mathcal{F}_{\mathbf{S}}$) is low, it indicates the presence of loose logical connections or errors within the reasoning,
in such cases, the final answer, even if correct, is more likely a byproduct of chance or internal bias than of logical necessity. Conversely, a low consistency score ($\mathcal{C}_{\mathbf{S}}$) indicates that the reasoning, however internally coherent, ultimately fails to solve the task. 
To bridge these two aspects, we define a reliable score $\mathcal{R}_{\mathbf{S}}$  by scaling observed correctness with reasoning faithfulness, thereby ensuring  that the final score reflects only outcomes grounded in logical integrity:
\vspace{-0.5em}
\begin{equation}
\nonumber
    \begin{aligned}
        &\mathcal{R}_{\mathbf{S}} = \mathcal{F}_{\mathbf{S}} \cdot \mathcal{C}_{\mathbf{S}} 
        \approx  \frac{1}{N(L-1)}  \sum_{i=1}^{L-1} \sum_{n=1}^{N} \\& \mathbf{1}\!\left[
X_i^{(\emptyset)} \succ X_i^{(\mathcal{E})}
\right] \cdot P(J^{(n)}=1 \mid q, \mathbf{S}) \cdot \mathbf{1}[J^{(n)} = 1],
    \end{aligned}
\end{equation}
where $L$ denotes the total number of steps in $\mathbf{S}$, $N$ is the number of answer-sampling trials, $X_i = (\mathbf{S}_{\leq i}, \mathbf{S}_{> i})$ represents the pair of reasoning segments split at position $i$, and $X_i^{(\emptyset)}=(\mathbf{S}_{\le i},\mathbf{S}_{>i}), X_i^{(\mathcal{E})}=(\mathbf{S}_{\le i},\mathbf{S}_{>i}^{\prime\,(\mathcal{E})}).$ Here, $\mathcal{E}$ denotes the perturbation setting used to generate the counterfactual continuation associated with split point $i$.

\begin{algorithm}[t!]
\footnotesize
\caption{Select the Trustworthy CoT via FACT-E (Lightweight)}
\label{alg:cot_causal_rerank_light}
\begin{algorithmic}[1]
\Require 
    Candidate set $\mathcal{S}=\{(\mathbf{S}^{(j)},a^{(j)})\}_{j=1}^{K}$;
    query $q$;
    language model $\mathcal{M}$;
    perturbation set $\mathcal{E}$;
    the number of sampling trials $N$
\Ensure 
    Optimal CoT $\mathbf{S}_{\mathrm{opt}}$,
    answer $a_{\mathrm{opt}}$,
    score $\mathcal{R}_{\max}$

\State $\mathrm{SC} \gets \emptyset$

\For{each candidate $(\mathbf{S}^{(j)},a^{(j)}) \in \mathcal{S}$}

    \State \textcolor{purple}{// \textbf{Step 1}: CoT-to-Answer Consistency}
    \State $\mathcal{C}_{\mathbf{S}^{(j)}} \gets \frac{1}{N}\sum_{n=1}^{N} P(J^{(n)}=1 \mid q,\mathbf{S}^{(j)}) \cdot \mathbf{1}[J^{(n)}=1]$
    \If{$\mathcal{C}_{\mathbf{S}^{(j)}} = 0$}
        \State \textbf{continue}
    \EndIf

    \State \textcolor{purple}{// \textbf{Step 2}: Intra-Chain Faithfulness}
    \State $L_j \gets |\mathbf{S}^{(j)}|$
    \State Sample $\mathcal{T}^{(j)} \subseteq \{1,\ldots,L_j-1\}$ such that $|\mathcal{T}^{(j)}|=\min(N,L_j-1)$
    \For{each $t \in \mathcal{T}^{(j)}$}
        \State $\mathbf{S}_{>t}^{\prime\,(e)} \gets \mathcal{M}(\cdot \mid \mathrm{InjectNoise}(\mathbf{S}_{\le t}^{(j)}, e)), \ \forall e \in \mathcal{E}$
        \State $P_{\mathrm{faith}}^{(j,t)} \gets \mathbb{E}_{e \in \mathcal{E}} \left[
        \mathbf{1}\!\left[
        (\mathbf{S}_{\le t}^{(j)},\mathbf{S}_{>t}^{(j)})
        \succ
        (\mathbf{S}_{\le t}^{(j)},\mathbf{S}_{>t}^{\prime\,(e)})
        \right]\right]$
    \EndFor
    \State $\mathcal{F}_{\mathbf{S}^{(j)}} \gets \frac{1}{|\mathcal{T}^{(j)}|}\sum_{t\in\mathcal{T}^{(j)}} P_{\mathrm{faith}}^{(j,t)}$

    \State \textcolor{purple}{// \textbf{Step 3}: FACT-E score}
    \State $\mathcal{R}_{\mathbf{S}^{(j)}} \gets \mathcal{C}_{\mathbf{S}^{(j)}} \cdot \mathcal{F}_{\mathbf{S}^{(j)}}$
    \State $\mathrm{SC} \gets \mathrm{SC} \cup \{(\mathbf{S}^{(j)},a^{(j)},\mathcal{R}_{\mathbf{S}^{(j)}})\}$
\EndFor

\State $(\mathbf{S}_{\mathrm{opt}},a_{\mathrm{opt}},\mathcal{R}_{\max}) \gets \arg\max_{(\mathbf{S},a,\mathcal{R}_{\mathbf{S}})\in \mathrm{SC}} \mathcal{R}_{\mathbf{S}}$
\State \Return $\mathbf{S}_{\mathrm{opt}},a_{\mathrm{opt}},\mathcal{R}_{\max}$
\end{algorithmic}
\label{algorithm1}
\end{algorithm}

\subsection{Algorithms} 

Evaluating $\mathcal{F}_{\mathbf{S}}$ at every step in a long reasoning chain can incur substantial computational overhead. We thus adopt a lightweight estimation strategy for $\mathcal{R}_{\mathbf{S}}$ using a fixed-checkpoint approach, where the number of checkpoints is set to $N$ to match the number of sampling trials for estimating $\mathcal{C}_{\mathbf{S}}$. Algorithm~\ref{alg:cot_causal_rerank_light} outlines the overall selection procedure. Specifically, for each CoT candidate, we first estimate its CoT-to-Answer Consistency ($\mathcal{C}_{\mathbf{S}}$). Candidates with zero consistency are discarded. 
For the remaining candidates, we sample $N$ random intermediate positions to estimate Intra-Chain Faithfulness. 
The optimal reasoning trace $\mathbf{S}_{\text{opt}}$ is then selected by jointly maximizing answer correctness and intermediate steps faithfulness. 
The standard (non-lightweight) version of the algorithm is provided in Algorithm~\ref{algorithm2}, and task-specific settings for error injection are detailed in Appendix~\ref{noise_inject}.

\definecolor{lightpurple}{RGB}{210, 190, 240} 
\definecolor{tablegray}{RGB}{250, 250, 250} 
\setlength{\tabcolsep}{4pt} 
\renewcommand{\arraystretch}{1.15} 
\begin{table*}[ht!]
\footnotesize 
\centering
\begin{tabular}{@{}lccc|@{\hspace{0.2em}}ccc@{}}
\toprule
\multicolumn{4}{c}{\textbf{Gpt-4o-mini}} & \multicolumn{3}{c}{\textbf{DeepSeek-V3}} \\
\cmidrule(r){1-4} \cmidrule(l){5-7}
\textbf{Method} & {Math-500} & {CommonsenseQA} & {GSM-8K} &
{Math-500} & {CommonsenseQA} & {GSM-8K} \\
\midrule
\textsc{CoT} & 78.69\textsubscript{\scriptsize(-4.253\%)} & 83.00\textsubscript{\scriptsize(+0.029\%)} & 92.40\textsubscript{\scriptsize(-0.154\%)} & 
85.52\textsubscript{\scriptsize(-4.587\%)} & 83.80\textsubscript{\scriptsize(-1.671\%)} & 96.00\textsubscript{\scriptsize(-0.243\%)} \\
\textsc{Denoise} & 83.06\textsubscript{\scriptsize(+0.117\%)} & 83.30\textsubscript{\scriptsize(+0.329\%)} & 92.20\textsubscript{\scriptsize(-0.354\%)} & 
84.70\textsubscript{\scriptsize(-5.407\%)} & 84.60\textsubscript{\scriptsize(-0.871\%)} & 95.40\textsubscript{\scriptsize(-0.843\%)} \\
\textsc{Polish} & 83.33\textsubscript{\scriptsize(+0.387\%)} & 80.00\textsubscript{\scriptsize(-2.971\%)} & 92.60\textsubscript{\scriptsize(+0.046\%)} & 
\underline{94.80}\textsubscript{\scriptsize(+4.693\%)} & 85.20\textsubscript{\scriptsize(-0.271\%)} & 95.60\textsubscript{\scriptsize(-0.643\%)} \\
\textsc{Refelect} & 80.60\textsubscript{\scriptsize(-2.343\%)} & 83.20\textsubscript{\scriptsize(+0.229\%)} & 92.10\textsubscript{\scriptsize(-0.454\%)} & 
87.43\textsubscript{\scriptsize(-2.677\%)} & \underline{85.80}\textsubscript{\scriptsize(+0.329\%)} & 95.80\textsubscript{\scriptsize(-0.443\%)} \\
\textsc{Consistency} & 82.79\textsubscript{\scriptsize(-0.153\%)} & 82.10\textsubscript{\scriptsize(-0.871\%)} & \underline{92.80}\textsubscript{\scriptsize(+0.246\%)} & 
93.17\textsubscript{\scriptsize(+3.063\%)} & 85.00\textsubscript{\scriptsize(-0.471\%)} & {96.40}\textsubscript{\scriptsize(+0.157\%)} \\
\textsc{FACT-E}\textsubscript{\tiny(Lightweight)} & \underline{85.52}\textsubscript{\scriptsize(+2.577\%)} & \textbf{85.20}\textsubscript{\scriptsize(+2.229\%)} & \textbf{93.98}\textsubscript{\scriptsize(+1.426\%)} & 
90.32\textsubscript{\scriptsize(+0.213\%)} & \textbf{89.00}\textsubscript{\scriptsize(+3.529\%)} & \underline{97.20}\textsubscript{\scriptsize(+0.957\%)} \\
\textsc{FACT-E}\textsubscript{\tiny(Standard)} & \textbf{86.61}\textsubscript{\scriptsize(+3.667\%)} & \underline{84.00}\textsubscript{\scriptsize(+1.029\%)} & 91.80\textsubscript{\scriptsize(-0.754\%)} & 
\textbf{94.81}\textsubscript{\scriptsize(+4.703\%)} & 84.90\textsubscript{\scriptsize(-0.571\%)} & \textbf{97.30}\textsubscript{\scriptsize(+1.057\%)} \\\hline
\multicolumn{4}{c}{\textbf{Qwen3}} & \multicolumn{3}{c}{\textbf{ChatGPT}} \\
\cmidrule(r){1-4} \cmidrule(l){5-7}
\textsc{CoT} & \underline{93.17}\textsubscript{\scriptsize(+0.154\%)} & 83.00\textsubscript{\scriptsize(-0.257\%)} & \underline{93.20}\textsubscript{\scriptsize(+0.171\%)} & 
\underline{52.18}\textsubscript{\scriptsize(+3.430\%)} & 61.60\textsubscript{\scriptsize(-7.471\%)} & 77.60\textsubscript{\scriptsize(-0.800\%)} \\
\textsc{Denoise} & 92.08\textsubscript{\scriptsize(-0.936\%)} & \underline{83.60}\textsubscript{\scriptsize(+0.343\%)} & \underline{93.20}\textsubscript{\scriptsize(+0.171\%)} & 
42.62\textsubscript{\scriptsize(-6.130\%)} & \underline{72.20}\textsubscript{\scriptsize(+3.129\%)} & 76.80\textsubscript{\scriptsize(-1.600\%)} \\
\textsc{Polish} & 92.90\textsubscript{\scriptsize(-0.116\%)} & 81.60\textsubscript{\scriptsize(-1.657\%)} & 92.40\textsubscript{\scriptsize(-0.629\%)} & 
41.00\textsubscript{\scriptsize(-7.750\%)} & 63.60\textsubscript{\scriptsize(-5.471\%)} & 77.00\textsubscript{\scriptsize(-1.400\%)} \\
\textsc{Refelect} & 92.90\textsubscript{\scriptsize(-0.116\%)} & 82.60\textsubscript{\scriptsize(-0.657\%)} & 92.60\textsubscript{\scriptsize(-0.429\%)} & 
48.09\textsubscript{\scriptsize(-0.660\%)} & 69.60\textsubscript{\scriptsize(+0.529\%)} & 78.40\textsubscript{\scriptsize(0.000\%)} \\
\textsc{Consistency} & 92.90\textsubscript{\scriptsize(-0.116\%)} & 82.80\textsubscript{\scriptsize(-0.457\%)} & \underline{93.20}\textsubscript{\scriptsize(+0.171\%)} & 
51.09\textsubscript{\scriptsize(+2.340\%)} & 71.60\textsubscript{\scriptsize(+2.529\%)} & \underline{80.60}\textsubscript{\scriptsize(+2.200\%)} \\
\textsc{FACT-E}\textsubscript{\tiny(Lightweight)} & \textbf{94.26}\textsubscript{\scriptsize(+1.244\%)} & \underline{83.60}\textsubscript{\scriptsize(+0.343\%)} & \underline{93.20}\textsubscript{\scriptsize(+0.171\%)} & 
\underline{52.18}\textsubscript{\scriptsize(+3.430\%)} & \textbf{72.70}\textsubscript{\scriptsize(+3.629\%)} & \textbf{81.40}\textsubscript{\scriptsize(+3.000\%)} \\
\textsc{FACT-E}\textsubscript{\tiny(Standard)} & 92.90\textsubscript{\scriptsize(-0.116\%)} & \textbf{85.60}\textsubscript{\scriptsize(+2.343\%)} & \textbf{93.40}\textsubscript{\scriptsize(+0.371\%)} & 
\textbf{54.09}\textsubscript{\scriptsize(+5.340\%)} & \underline{72.20}\textsubscript{\scriptsize(+3.129\%)} & 77.00\textsubscript{\scriptsize(-1.400\%)} \\
\bottomrule
\end{tabular}
\caption{Accuracy comparison (\%) across three benchmarks for four LLMs. Values in parentheses denote the relative change with respect to the average performance of all methods for each model-dataset pair. Best results are shown in \textbf{Bold}, second-best results are \underline{underlined}.}
\label{tab:summary_results_final}
\end{table*}

\section{Experimental Setup}

In this section, we evaluate FACT-E through the following research questions:
(1) \textbf{RQ1}: Can FACT-E identify trustworthy reasoning trajectories to improve answer accuracy?
(2) \textbf{RQ2}: Can the selected CoTs serve as superior exemplars for in-context learning?
(3) \textbf{RQ3}: Can FACT-E effectively detect and filter rationale noise to safeguard performance?

\paragraph{Evaluation for RQ1 and RQ2.}
We conduct two experiments for RQ1 and RQ2. \textit{(1) Trustworthy reasoning trajectory selection.}
For each test set $\mathcal{Q}=\{(q_t,a_t)\}_{t=1}^{|\mathcal{Q}|}, $
we generate $K$ candidate CoT-answer pairs by sampling: $\mathcal{S}(q_t)=\{(\mathbf{S}^{(j)}_t,a^{(j)}_t)\}_{j=1}^{K}.
$ FACT-E selects the best reasoning trajectory $(\mathbf{S}^{*}_t,a^{*}_t)$ with highest $\mathcal{R}_{\mathbf{S}}.
$We report answer accuracy as
$\mathrm{Acc}(\mathcal{Q})=\frac{1}{|\mathcal{Q}|}\sum_{t=1}^{|\mathcal{Q}|}\mathbf{1}\!\left[a^{*}_t=a_t\right].$ \textit{(2) In-context learning (ICL) evaluation.}
We further evaluate whether the automatically selected chains serve as higher-quality ICL exemplars.
For an exemplar size $E \in \{5,10,15\}$, we construct a prompt set
$\mathcal{P}_E=[(q_1,\mathbf{S}^{*}_1,a_1),\ldots,(q_E,\mathbf{S}^{*}_E,a_E)].$
The ICL accuracy is defined as $\mathrm{Acc}_{\mathrm{ICL}}(\mathcal{P}_E,\mathcal{Q})=\frac{1}{|\mathcal{Q}|}
\sum_{(q_t,a_t)\in Q}
\mathbf{1}\!\left[\mathcal{M}(\mathcal{P}_E,q_t)=a_t\right],$ where $\mathcal{M}(\mathcal{P}_E,q_t)$ denotes the model prediction conditioned on the exemplars. To evaluate FACT-E across domains and difficulty levels, we use public datasets: GSM8K~\citep{cobbe2021training} and MATH-500~\citep{hendrycks2021measuring} for mathematical reasoning, together with CommonsenseQA~\citep{talmor2019commonsenseqa}.

\paragraph{Evaluation for RQ3.}
Let
$
\mathcal{Q}=\{(q_t,a_t)\}_{t=1}^{|\mathcal{Q}|}
$
be the test set. For each test query \(q_t\), we are given a noisy demonstration prompt
$
\mathcal{P}_{\mathrm{noise}}^{(t)}
=
\{(q_i,\mathbf{S}^{(i)},a_i)\}_{i=1}^{K_t},
$
where the rationales are taken directly from the benchmark and may already contain noisy intermediate steps. The number of noisy rationales in each prompt is not fixed and varies across evaluation settings. Given a filtering method $\mathcal{M}$, we define its answering accuracy under noisy demonstrations as
$
\mathrm{Acc}_{\mathrm{noise}}(\mathcal{Q},\mathcal{P}_{\mathrm{noise}})
=
\frac{1}{|\mathcal{Q}|}
\sum_{t=1}^{|\mathcal{Q}|}
\mathbf{1}\!\left[
\mathcal{M}\!\left(\mathcal{P}_{\mathrm{noise}}^{(t)}, q_t\right)=a_t
\right].
$
We evaluate robustness on NoRa-Math and NoRa-Commonsense~\citep{zhou2024can}. The detailed procedure is shown in Algorithm~\ref{alg:denoise}.

\definecolor{bestpurple}{RGB}{190, 200, 255}
\definecolor{secondpurple}{RGB}{225, 235, 255}

\newcommand{\deltatext}[1]{\scriptsize\textcolor{black!75}{(#1)}}

\newcommand{\score}[2]{\makecell[c]{#1\\[-2pt]\deltatext{#2}}}
\newcommand{\bestscore}[2]{\cellcolor{bestpurple}\makecell[c]{\textbf{#1}\\[-2pt]\deltatext{#2}}}
\newcommand{\secondscore}[2]{\cellcolor{secondpurple}\makecell[c]{\underline{#1}\\[-2pt]\deltatext{#2}}}
\newcommand{\bestbase}[1]{\cellcolor{bestpurple}\textbf{#1}}
\newcommand{\secondbase}[1]{\cellcolor{secondpurple}\underline{#1}}

\begin{table*}[t]
\centering
\footnotesize
\renewcommand{\arraystretch}{1.15}
\setlength{\tabcolsep}{2pt}

\begin{adjustbox}{max width=\textwidth}
\begin{tabular}{@{}lcccccccccccc@{}}
\toprule
\multirow{2}{*}{Method}
& \multicolumn{4}{c}{MATH-500}
& \multicolumn{4}{c}{CommonsenseQA}
& \multicolumn{4}{c}{GSM-8K} \\
\cmidrule(lr){2-5} \cmidrule(lr){6-9} \cmidrule(lr){10-13}
& 4o-mini & DeepSeek-V3 & Qwen3 & ChatGPT
& 4o-mini & DeepSeek-V3 & Qwen3 & ChatGPT
& 4o-mini & DeepSeek-V3 & Qwen3 & ChatGPT \\
\midrule

\textsc{Base}
& 82.79 & 92.35 & 90.16 & 46.99
& 83.00 & 86.00 & 81.40 & 49.60
& \bestbase{93.20} & \secondbase{94.60} & 93.20 & 74.20 \\

\textsc{Denoise}
& \score{82.65}{-0.14} & \score{92.62}{+0.27} & \score{85.79}{-4.37} & \score{47.54}{+0.55}
& \score{82.80}{-0.20} & \bestscore{86.60}{+0.60} & \score{81.00}{-0.40} & \secondscore{51.60}{+2.00}
& \secondscore{93.00}{-0.20} & \score{93.60}{-1.00} & \secondscore{93.80}{+0.60} & \secondscore{77.00}{+2.80} \\

\textsc{Polish}
& \secondscore{83.42}{+0.63} & \bestscore{94.81}{+2.46} & \secondscore{92.08}{+1.92} & \score{49.73}{+2.74}
& \score{81.80}{-1.20} & \score{84.80}{-1.20} & \score{81.00}{-0.40} & \score{52.22}{+2.62}
& \score{91.60}{-1.60} & \score{91.60}{-3.00} & \score{93.60}{+0.40} & \score{71.80}{-2.40} \\

\textsc{Reflect}
& \score{83.35}{+0.56} & \secondscore{93.44}{+1.09} & \score{88.25}{-1.91} & \secondscore{51.64}{+4.65}
& \score{81.80}{-1.20} & \score{86.00}{+0.00} & \score{80.80}{-0.60} & \score{53.20}{+3.60}
& \score{91.00}{-2.20} & \score{93.80}{-0.80} & \score{92.80}{-0.40} & \score{76.20}{+2.00} \\

\textsc{Consist.}
& \score{82.89}{+0.10} & \score{90.98}{-1.37} & \score{91.26}{+1.10} & \score{46.99}{+0.00}
& \secondscore{83.80}{+0.80} & \score{82.00}{-4.00} & \secondscore{81.80}{+0.40} & \score{55.60}{+6.00}
& \score{92.00}{-1.20} & \score{94.40}{-0.20} & \score{92.40}{-0.80} & \score{71.20}{-3.00} \\

\textsc{FACT-E}
& \bestscore{85.52}{+2.73} & \secondscore{93.44}{+1.09} & \bestscore{92.62}{+2.46} & \bestscore{53.28}{+6.29}
& \bestscore{84.00}{+1.00} & \secondscore{86.20}{+0.20} & \bestscore{82.40}{+1.00} & \bestscore{63.40}{+13.80}
& \score{92.20}{-1.00} & \bestscore{95.00}{+0.40} & \bestscore{94.80}{+1.60} & \bestscore{77.40}{+3.20} \\

\bottomrule
\end{tabular}
\end{adjustbox}

\caption{
Accuracy (\%) of in-context learning with five demonstration examples ($E = 5$).
Best results are shown in \textbf{bold} with darker shading, and second-best results are \underline{underlined} with lighter shading.
Values in parentheses denote the absolute change relative to \textsc{Base}.
}
\label{tab:results_icl}
\end{table*}

\paragraph{LLM backbones and baseline methods.} 
We conduct evaluations using four LLM backbones ranging from open-source to closed-source models, including DeepSeek-V3, Qwen3-14B, Gpt-4o-mini, and ChatGPT (Gpt-3.5-turbo). For all models, we set the temperature parameter $\tau$ to 0 and the number of sampling trials to $N=3$, and we conduct experiments in other $N$ settings which are reported in \S\ref{results}. To ensure stable estimates, we evaluate 500 questions per task and repeat each experiment three times. We compare our method against five representative baselines. CoT~\citep{wei2022chain}, \textsc{Polish}~\citep{xi2023self} and \textsc{Reflect}~\citep{kadavath2022language} fall under the self-correction paradigm, which aims to improve generation quality through prompt reformulation and iterative reflection. \textsc{Denoise}~\citep{zhang2023certified} adopts a mask-reconstruction strategy that requires the model to recover masked content, while \textsc{Consistency}~\citep{wang2022self} aims to aggregate multiple sampled outputs for improving robustness. These baselines represent mainstream self-improvement and denoising approaches. The detailed descriptions of these baselines are provided in Appendix~\ref{appendix: baselines}. 

\section{Experimental Results}\label{results}

\paragraph{FACT-E can select trustworthy chains and improve answer accuracy.} 
As shown in Table~\ref{tab:summary_results_final}, FACT-E exhibits clear advantages across 12 experimental configurations involving four LLMs and three benchmarks. The standard version of FACT-E achieves the best or second-best performance in 8 out of 12 cases. In particular, FACT-E (standard) shows substantial improvements on the Math-500 benchmark, outperforming the average baseline by 5.340\% on ChatGPT and 4.703\% on DeepSeek-V3. A closer inspection reveals that FACT-E effectively identifies trustworthy CoT trajectories, achieving 54.09\% accuracy on MATH-500, compared to self-correction baselines such as POLISH (41.00\%) and DENOISE (42.62\%). Moreover, the lightweight variant of FACT-E remains highly competitive: across the same 12 configurations, FACT-E (lightweight) attains the best result in 6 cases and the second-best in 5 cases. This indicates that even with stochastic checkpoint sampling, FACT-E preserves strong discriminative power while significantly reducing computational overhead. 
\paragraph{Selected trustworthy CoTs enhance ICL.}\label{section:in-context-learning} 
As reported in Table~\ref{tab:results_icl}, FACT-E achieves the best performance in 8 out of 12 task configurations, with particularly notable gains on ChatGPT (e.g., from 49.60\% to 63.40\% on CommonsenseQA and from 46.99\% to 53.28\% on MATH-500). Compared with competing baselines, FACT-E exhibits substantially higher stability across different models and benchmarks. For example, while \textsc{POLISH} performs competitively on MATH using DeepSeek-V3 and Qwen3, its performance degrades using ChatGPT (on MATH) and even falls below the Base method using Gpt-4o-mini (on CommonsenseQA). These observations suggest that methods relying solely on self-assessment are sensitive to the underlying model's capabilities, whereas FACT-E remains robust across architectures.
\definecolor{casestudy}{RGB}{207 207 207}
\definecolor{wrong}{RGB}{205 50 120}  
\definecolor{right}{RGB}{58 95 205}
\newcommand{\cmark}{\ding{51}}  
\newcommand{\xmark}{\ding{55}}
\begin{table*}[t!]
\centering
\tiny
\label{tab:qa_evaluation}
\begin{tabular}{p{0.1cm}p{1.3cm}p{13cm}p{0.1cm}}
\toprule
\textbf{\#} & \textbf{Answer} & \textbf{CoT (Chain of Thought)} & \textbf{Score} \\
\midrule

\rowcolor{casestudy}
\multicolumn{4}{p{\textwidth}}{%
Question: Suppose $\sin D = 0.7$ in the diagram below. What is $DE$? [asy] pair D,E,F; F = (0,0); D = (sqrt(51),7); E = (0,7); draw(D--E--F--D); draw(rightanglemark(D,E,F,15)); label("$D$",D,NE); label("$E$",E,NW); label("$F$",F,SW); label("$7$",(E+F)/2,W); [/asy]} \\

1 & $\sqrt{51}$ \textcolor{green}{\Checkmark} & 
We are given that $\sin D = 0.7$ and we need to find the length of $DE$. In the right triangle $DEF$, we know that $EF$ is the vertical side and has a length of $7$. The sine of angle $D$ is defined as the ratio of the length of the opposite side to the hypotenuse. Therefore, we can express this relationship as:
$\sin D = \frac{EF}{DF}$. Substituting the known values, we have: $0.7 = \frac{7}{DF}$
To find $DF$, we can rearrange the equation: $DF = \frac{7}{0.7}$. Calculating this gives:$DF = 10$Now, we can use the Pythagorean theorem to find $DE$. In triangle $DEF$:$DF^2 = DE^2 + EF^2$
Substituting the known values:$10^2 = DE^2 + 7^2$This simplifies to:$100 = DE^2 + 49$Now, isolating $DE^2$:$DE^2 = 100 - 49 = 51$ Taking the square root gives:$DE = \sqrt{51}$ Thus, the length of $DE$ is $\sqrt{51}$.
& 0.998 \\

2 & $10$ \textcolor{red}{\XSolidBrush} &
We are given that $\sin D = 0.7$ and we need to find the length of $DE$. In the right triangle $DEF$, we know that $EF$ is the opposite side to angle $D$ and $DE$ is the hypotenuse. The sine function is defined as the ratio of the length of the opposite side to the hypotenuse. Therefore, we can express this relationship as:
\textcolor{wrong}{$\sin D = \frac{EF}{DE}$
From the diagram, we see that $EF = 7$. Substituting the known values into the sine equation gives us:
$0.7 = \frac{7}{DE}$}
To find $DE$, we can rearrange this equation:
$DE = \frac{7}{0.7}$
Calculating this gives:
$DE = \frac{7}{0.7} = 10$
Thus, the length of $DE$ is $10$.
& 0 \\

\midrule

\rowcolor{casestudy}
\multicolumn{4}{p{\textwidth}}{%
Question: What is $1^{(2^{235423523})}$?} \\

1 & $1$ \textcolor{green}{\Checkmark} &
To solve the expression $1^{(2^{235423523})}$, we need to understand the properties of exponents. The base of the exponent is 1. According to the rules of exponents, any number raised to any power is equal to 1, as long as the base is 1. Therefore, regardless of the value of the exponent, $1$ raised to any power will always equal $1$.
& 0.999 \\

2 & $1$ \textcolor{green}{\Checkmark}&
The expression $1^{(2^{235423523})}$ involves raising 1 to the power of $2^{235423523}$. \textcolor{wrong}{The key property of exponentiation to note here is that any number raised to the power of 0 is 1}, and any number raised to any power (including very large numbers) is still 1 if the base is 1. Therefore, regardless of the exponent, since the base is 1, the result will always be 1.
& 0.596 \\
\bottomrule
\end{tabular}
\caption{Question-answer evaluation results. Flawed portions of the CoT are highlighted in \textcolor{wrong}{red}. Correct and incorrect answers are marked with \textcolor{green}{\Checkmark} and \textcolor{red}{\XSolidBrush}, respectively.}
\label{case_study}
\end{table*}
\begin{figure}[t!]
    \centering
    \includegraphics[width=0.98\linewidth]{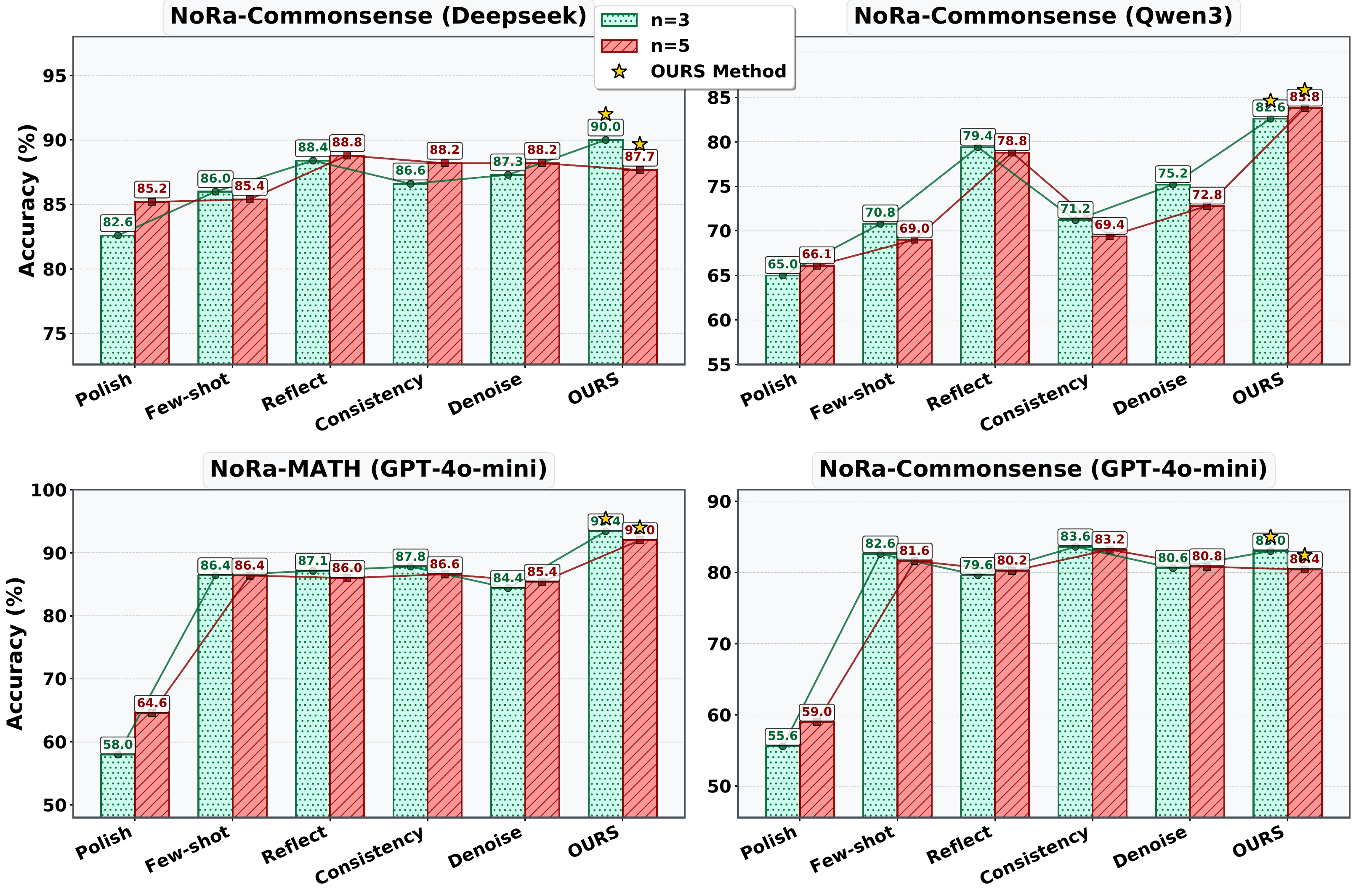}
    \caption{Performance (\%) on noisy-rationale benchmarks. Results are shown for Qwen3 and DeepSeek-V3 on NoRa-Commonsense, and Gpt-4o-mini on NoRa-Math/Commonsense, under in-context learning with different numbers of noisy rationale demonstrations.}

    \label{fig:denoise}

\end{figure}

\paragraph{FACT-E for noisy CoT detection.}
To evaluate robustness against rationale noise, we test FACT-E on NoRa prompting settings in which each prompt contains a variable number of noisy rationales. In particular, we consider settings with 1--3 and 1--5 noisy rationales per prompt (see Algorithm~\ref{alg:denoise}), simulating realistic cases where demonstration chains contain incorrect or weak intermediate reasoning steps. As shown in Figure~\ref{fig:denoise}, FACT-E consistently achieves top accuracy on NoRa-Commonsense with Qwen3 and DeepSeek-V3. Although GPT-4o-mini is slightly less effective on NoRa-Commonsense, it achieves leading performance on NoRa-Math. These results underscore the effectiveness of FACT-E in identifying and filtering unreliable reasoning demonstrations.

\paragraph{Case study.}
As illustrated in Table~\ref{case_study}, FACT-E effectively distinguishes CoT candidates that contain flawed reasoning processes. In the first example, FACT-E differentiates between two reasoning paths: although CoT~1 yields the correct final answer, CoT~2 produces an incorrect result due to the erroneous trigonometric derivation ``$\text{sin} D = EF/DE$''. FACT-E assigns this chain a score of 0, correctly indicating that it cannot reliably support the correct conclusion. Additional examples further demonstrate FACT-E's ability to identify reasoning paths whose intermediate steps lack causal validity despite arriving at the correct answer. In the second case, the highlighted red text corresponds to redundant reasoning where the transition across ``and'' lacks a rigorous causal dependency, resulting in a lower score of 0.596. These fine-grained evaluations show that FACT-E provides a more nuanced characterization of reasoning quality beyond final-answer correctness. Additional case studies are provided in Table~\ref{app:cases}.

\begin{figure}[t]
    \centering
    \includegraphics[width=1\linewidth]{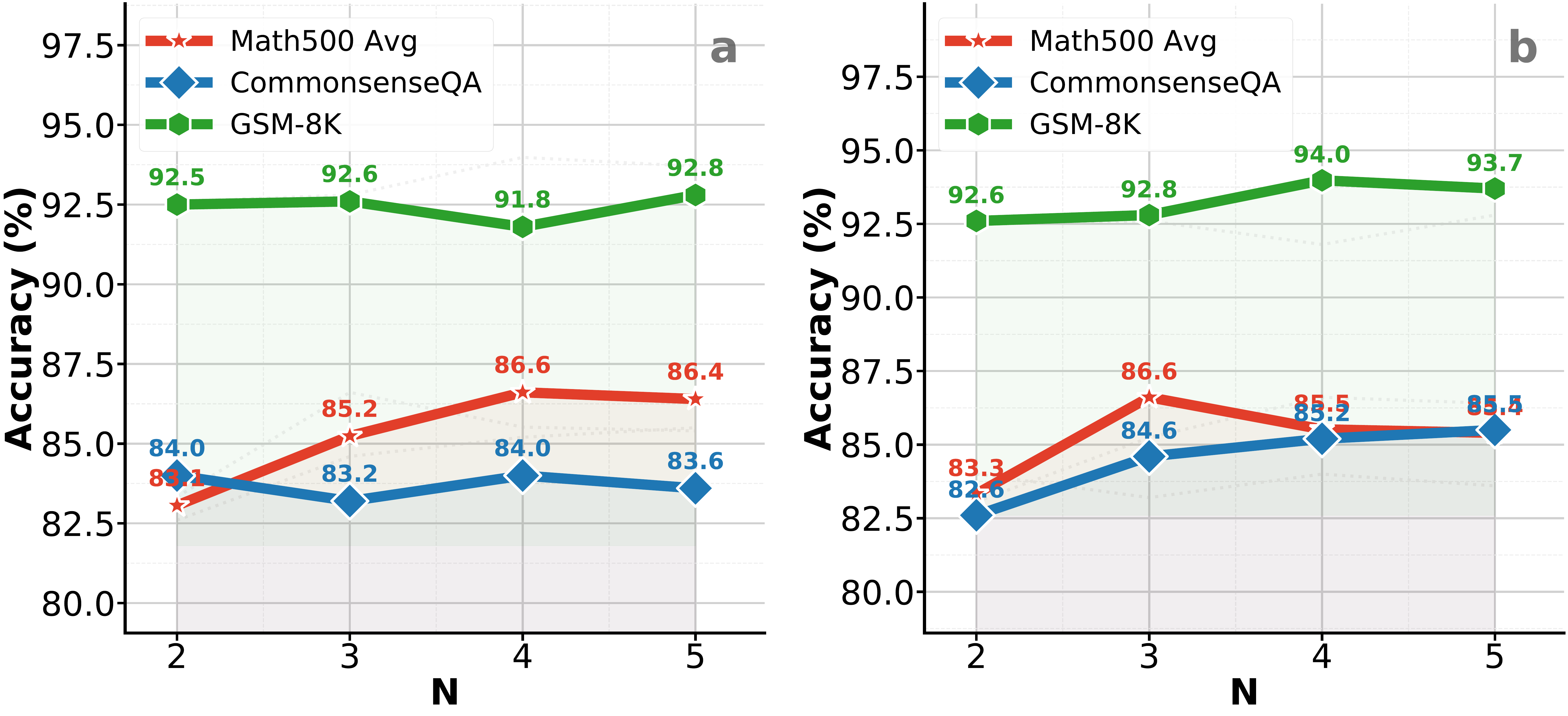}
    \vspace{-1.5em}
    \caption{Performance (\%) of lightweight FACT-E (a) and standard FACT-E (b) varying the number of sampling trials ($N$) on three benchmarks.}

    \vspace{-1.5em}

    \label{fig:placeholder}
\end{figure}

\paragraph{Analysis of the number of sampling trials ($N$).} 
Experimental results within Gpt-4o-mini across datasets indicate that accuracy generally improves with more sampling trials, although the gains are not strictly monotonic. As shown in Figures~\ref{fig:placeholder}(a) (lightweight) and~\ref{fig:placeholder}(b) (standard), accuracy often increases substantially between the second and third trials, followed by saturation or minor fluctuations at four or five trials. This suggests that three trials typically capture most of the performance gains, while additional iterations yield diminishing returns.

\section{Related Work}
Previous work on improving CoT reasoning~\citep{wei2023chainofthoughtpromptingelicitsreasoning,kojima2023largelanguagemodelszeroshot} has mainly focused on prompt design~\citep{zhou2022least,wang2023selfconsistencyimproveschainthought}; structured search frameworks~\citep{sel2023algorithm,yu2023thought} such as Tree-of-Thought~\citep{yao2023tree} and Graph-of-Thought~\citep{Besta_Blach_Kubicek_Gerstenberger_Podstawski_Gianinazzi_Gajda_Lehmann_Niewiadomski_Nyczyk_Hoefler_2024}; and fine-tuned language models in specific domains~\citep{jiang-etal-2023-llmlingua,sun2025explainable,wang2025think,huang2026semanticspaceexplorationexploitationrlvr}. These methods aim to improve final-answer accuracy, often treating accuracy gains as indirect evidence of better reasoning. However, they do not directly address a key practical question~\citep{shen2025faithcot}: given a specific query and a generated CoT, how can the quality of that reasoning trace be reliably assessed?

Existing approaches to CoT evaluation broadly fall into two categories~\citep{wei2022chain,kojima2023largelanguagemodelszeroshot}. \textit{LLM-based assessment methods}, including self-correction, self-reflection, and self-refinement~\citep{kadavath2022languagemodelsmostlyknow,madaan2023self,xi2024selfpolishenhancereasoninglarge}, rely on the model’s own judgments to evaluate or improve its reasoning. While effective in some settings, these methods assume reliable self-evaluation and are therefore sensitive to model biases. \textit{Causality-based methods} attempt to assess reasoning quality by analyzing dependencies among intermediate steps, for example using Probability of Necessity and Sufficiency (PNS) or Average Causal Effect (ACE)~\citep{yu2025causalsufficiencynecessityimproves,fu2025unveilingcausalizingcotcausal}. However, these approaches depend heavily on LLM self-assessment, lack principled uncertainty quantification and face scalability limitations. In contrast, our work focuses on rigorously evaluating CoT reasoning by disentangling step-level faithfulness dependencies while explicitly addressing confounding effects (i.e., internal bias), providing a more reliable and scalable framework for LLM reasoning evaluation.

\section{Conclusion}
We address the inherent bias in LLM self-evaluation by introducing a causal framework based on Structural Causal Models, named FACT-E. By leveraging constructed noise as an instrumental variable to estimate the Average Causal Effect, our approach isolates the true causal influence of intermediate reasoning steps by effectively mitigating spurious correlations arising from internal model biases (e.g., self-affirmation bias), enabling a more reliable estimation of reasoning faithfulness.

\section*{Limitations}
Our approach, standard FACT-E, aims to assess all the steps of CoT. While it requires prompt LLMs multiple times, leading to higher inference costs compared to simpler prompting approaches. To mitigate the cost, we introduce lightweight FACT-E, which reduces the number of prompting requests while maintaining competitive performance. We compare the LLMs' overhead during inference using different strategies, shown in Tables~\ref{table:overhead}.

LLMs are inherently susceptible to a broad spectrum of cognitive biases~\citep{jiang2024peek,xiong2025measuring,zheng2023judging} and pre-trained data bias~\citep{jia2026unsupervised,wang2024word}. While it is impossible to account for every potential artifact, our framework specifically targets self-affirmation bias and shortcut bias, both of which significantly distort self-assessment tasks~\citep{xiong2025measuring,zheng2023judging}. Furthermore, our approach operates in a post-hoc manner; it selects from completed outputs instead of intervening in the CoT or reasoning generation process~\citep{luo2026gcot,cao2026diffcot,wu2026spark,shen2026double,xu2026selfcorrectingragenhancingfaithfulness,xu2026rcbsfmultiagentframeworkautomated,he2026ideainterpretableeditabledecisionmaking}. While cross-model or dynamic evaluation is common~\citep{li2025mtr}, it often introduces significant uninterpretable variables, such as inter-model sycophancy or shared parametric biases, which can lead to a false sense of consensus~\citep{du2023improving}. 

\section*{Ethics Statement}
All evaluations were conducted using open-source datasets. Our data sources are all from objective and neutral facts and do not contain any personal information and offensive comments directed at individuals or particular groups. Our study on mitigating bias in LLMs acknowledges the ethical implications of data-driven biases in AI, particularly their impact on performance. All experiments were conducted using publicly available datasets, and no human participants were involved.

\section*{Acknowledge}
This work is supported by the National Natural Science Foundation of China Young Scientists Fund (No. 62206233). MG was supported by the Australian Government through the ARC Discovery Projects (DP240102088).

\bibliography{custom}

\begin{thebibliography}{56}
\providecommand{\natexlab}[1]{#1}

\bibitem[{Besta et~al.(2024)Besta, Blach, Kubicek, Gerstenberger, Podstawski, Gianinazzi, Gajda, Lehmann, Niewiadomski, Nyczyk, and Hoefler}]{Besta_Blach_Kubicek_Gerstenberger_Podstawski_Gianinazzi_Gajda_Lehmann_Niewiadomski_Nyczyk_Hoefler_2024}
Maciej Besta, Nils Blach, Ales Kubicek, Robert Gerstenberger, Michal Podstawski, Lukas Gianinazzi, Joanna Gajda, Tomasz Lehmann, Hubert Niewiadomski, Piotr Nyczyk, and Torsten Hoefler. 2024.
\newblock \href {https://doi.org/10.1609/aaai.v38i16.29720} {Graph of thoughts: Solving elaborate problems with large language models}.
\newblock \emph{Proceedings of the AAAI Conference on Artificial Intelligence}, 38(16):17682--17690.

\bibitem[{Cao et~al.(2026)Cao, Lin, Gu, Luo, and Ma}]{cao2026diffcot}
Shidong Cao, Hongzhan Lin, Yuxuan Gu, Ziyang Luo, and Jing Ma. 2026.
\newblock Diffcot: Diffusion-styled chain-of-thought reasoning in llms.
\newblock \emph{arXiv preprint arXiv:2601.03559}.

\bibitem[{Cobbe et~al.(2021)Cobbe, Kosaraju, Bavarian, Chen, Jun, Kaiser, Plappert, Tworek, Hilton, Nakano et~al.}]{cobbe2021training}
Karl Cobbe, Vineet Kosaraju, Mohammad Bavarian, Mark Chen, Heewoo Jun, Lukasz Kaiser, Matthias Plappert, Jerry Tworek, Jacob Hilton, Reiichiro Nakano, and 1 others. 2021.
\newblock Training verifiers to solve math word problems.
\newblock \emph{arXiv preprint arXiv:2110.14168}.

\bibitem[{Cui et~al.(2024)Cui, He, Tang, He, Luo, Tang, and Xing}]{cui2024theoretical}
Yingqian Cui, Pengfei He, Xianfeng Tang, Qi~He, Chen Luo, Jiliang Tang, and Yue Xing. 2024.
\newblock A theoretical understanding of chain-of-thought: Coherent reasoning and error-aware demonstration.
\newblock \emph{arXiv preprint arXiv:2410.16540}.

\bibitem[{Du et~al.(2023)Du, Li, Torralba, Tenenbaum, and Mordatch}]{du2023improving}
Yilun Du, Shuang Li, Antonio Torralba, Joshua~B Tenenbaum, and Igor Mordatch. 2023.
\newblock Improving factuality and reasoning in language models through multiagent debate.
\newblock In \emph{Forty-first International Conference on Machine Learning}.

\bibitem[{Fu et~al.(2025{\natexlab{a}})Fu, Ding, Li, Li, Wei, and Chen}]{fu2025unveiling}
Jiarun Fu, Lizhong Ding, Hao Li, Pengqi Li, Qiuning Wei, and Xu~Chen. 2025{\natexlab{a}}.
\newblock Unveiling and causalizing cot: A causal pespective.
\newblock \emph{arXiv preprint arXiv:2502.18239}.

\bibitem[{Fu et~al.(2025{\natexlab{b}})Fu, Ding, Li, Li, Wei, and Chen}]{fu2025unveilingcausalizingcotcausal}
Jiarun Fu, Lizhong Ding, Hao Li, Pengqi Li, Qiuning Wei, and Xu~Chen. 2025{\natexlab{b}}.
\newblock \href {https://arxiv.org/abs/2502.18239} {Unveiling and causalizing cot: A causal pespective}.
\newblock \emph{Preprint}, arXiv:2502.18239.

\bibitem[{He et~al.(2026)He, Jiang, Wu, Huang, Wei, and Wang}]{he2026ideainterpretableeditabledecisionmaking}
Yanji He, Yuxin Jiang, Yiwen Wu, Bo~Huang, Jiaheng Wei, and Wei Wang. 2026.
\newblock \href {https://arxiv.org/abs/2604.12573} {Idea: An interpretable and editable decision-making framework for llms via verbal-to-numeric calibration}.
\newblock \emph{Preprint}, arXiv:2604.12573.

\bibitem[{Hendrycks et~al.(2021)Hendrycks, Burns, Kadavath, Arora, Basart, Tang, Song, and Steinhardt}]{hendrycks2021measuring}
Dan Hendrycks, Collin Burns, Saurav Kadavath, Akul Arora, Steven Basart, Eric Tang, Dawn Song, and Jacob Steinhardt. 2021.
\newblock Measuring mathematical problem solving with the math dataset.
\newblock \emph{arXiv preprint arXiv:2103.03874}.

\bibitem[{Huang et~al.(2026)Huang, Huang, Fan, He, Liang, Chen, Jiang, Khan, Jiang, and Wang}]{huang2026semanticspaceexplorationexploitationrlvr}
Fanding Huang, Guanbo Huang, Xiao Fan, Yi~He, Xiao Liang, Xiao Chen, Qinting Jiang, Faisal~Nadeem Khan, Jingyan Jiang, and Zhi Wang. 2026.
\newblock \href {https://arxiv.org/abs/2509.23808} {Semantic-space exploration and exploitation in rlvr for llm reasoning}.
\newblock \emph{Preprint}, arXiv:2509.23808.

\bibitem[{Huang et~al.(2023)Huang, Chen, Mishra, Zheng, Yu, Song, and Zhou}]{huang2023large}
Jie Huang, Xinyun Chen, Swaroop Mishra, Huaixiu~Steven Zheng, Adams~Wei Yu, Xinying Song, and Denny Zhou. 2023.
\newblock Large language models cannot self-correct reasoning yet.
\newblock \emph{arXiv preprint arXiv:2310.01798}.

\bibitem[{Hüyük et~al.(2025)Hüyük, Xu, Maasch, Nori, and González}]{hüyük2025reasoningelicitationlanguagemodels}
Alihan Hüyük, Xinnuo Xu, Jacqueline Maasch, Aditya~V. Nori, and Javier González. 2025.
\newblock \href {https://arxiv.org/abs/2410.03767} {Reasoning elicitation in language models via counterfactual feedback}.
\newblock \emph{Preprint}, arXiv:2410.03767.

\bibitem[{Jia et~al.(2026)Jia, Wu, Chen, Jing, Wang, Bu, and Wu}]{jia2026unsupervised}
Junhao Jia, Yueyi Wu, Huangwei Chen, Haodong Jing, Haishuai Wang, Jiajun Bu, and Lei Wu. 2026.
\newblock Unsupervised causal prototypical networks for de-biased interpretable dermoscopy diagnosis.
\newblock \emph{arXiv preprint arXiv:2602.23752}.

\bibitem[{Jiang et~al.(2024{\natexlab{a}})Jiang, Xie, Hao, Wang, Mallick, Su, Taylor, and Roth}]{jiang2024peek}
Bowen Jiang, Yangxinyu Xie, Zhuoqun Hao, Xiaomeng Wang, Tanwi Mallick, Weijie~J Su, Camillo~Jose Taylor, and Dan Roth. 2024{\natexlab{a}}.
\newblock A peek into token bias: Large language models are not yet genuine reasoners.
\newblock In \emph{Proceedings of the 2024 Conference on Empirical Methods in Natural Language Processing}, pages 4722--4756.

\bibitem[{Jiang et~al.(2024{\natexlab{b}})Jiang, Xie, Hao, Wang, Mallick, Su, Taylor, and Roth}]{jiang-etal-2024-peek}
Bowen Jiang, Yangxinyu Xie, Zhuoqun Hao, Xiaomeng Wang, Tanwi Mallick, Weijie~J Su, Camillo~Jose Taylor, and Dan Roth. 2024{\natexlab{b}}.
\newblock \href {https://doi.org/10.18653/v1/2024.emnlp-main.272} {A peek into token bias: Large language models are not yet genuine reasoners}.
\newblock In \emph{Proceedings of the 2024 Conference on Empirical Methods in Natural Language Processing}, pages 4722--4756, Miami, Florida, USA. Association for Computational Linguistics.

\bibitem[{Jiang et~al.(2023)Jiang, Wu, Lin, Yang, and Qiu}]{jiang-etal-2023-llmlingua}
Huiqiang Jiang, Qianhui Wu, Chin-Yew Lin, Yuqing Yang, and Lili Qiu. 2023.
\newblock \href {https://doi.org/10.18653/v1/2023.emnlp-main.825} {{LLML}ingua: Compressing prompts for accelerated inference of large language models}.
\newblock In \emph{Proceedings of the 2023 Conference on Empirical Methods in Natural Language Processing}, pages 13358--13376, Singapore. Association for Computational Linguistics.

\bibitem[{Kadavath et~al.(2022{\natexlab{a}})Kadavath, Conerly, Askell, Henighan, Drain, Perez, Schiefer, Hatfield-Dodds, DasSarma, Tran-Johnson, Johnston, El-Showk, Jones, Elhage, Hume, Chen, Bai, Bowman, Fort, Ganguli, Hernandez, Jacobson, Kernion, Kravec, Lovitt, Ndousse, Olsson, Ringer, Amodei, Brown, Clark, Joseph, Mann, McCandlish, Olah, and Kaplan}]{kadavath2022languagemodelsmostlyknow}
Saurav Kadavath, Tom Conerly, Amanda Askell, Tom Henighan, Dawn Drain, Ethan Perez, Nicholas Schiefer, Zac Hatfield-Dodds, Nova DasSarma, Eli Tran-Johnson, Scott Johnston, Sheer El-Showk, Andy Jones, Nelson Elhage, Tristan Hume, Anna Chen, Yuntao Bai, Sam Bowman, Stanislav Fort, and 17 others. 2022{\natexlab{a}}.
\newblock \href {https://arxiv.org/abs/2207.05221} {Language models (mostly) know what they know}.
\newblock \emph{Preprint}, arXiv:2207.05221.

\bibitem[{Kadavath et~al.(2022{\natexlab{b}})Kadavath, Conerly, Askell, Henighan, Drain, Perez, Schiefer, Hatfield-Dodds, DasSarma, Tran-Johnson et~al.}]{kadavath2022language}
Saurav Kadavath, Tom Conerly, Amanda Askell, Tom Henighan, Dawn Drain, Ethan Perez, Nicholas Schiefer, Zac Hatfield-Dodds, Nova DasSarma, Eli Tran-Johnson, and 1 others. 2022{\natexlab{b}}.
\newblock Language models (mostly) know what they know.
\newblock \emph{arXiv preprint arXiv:2207.05221}.

\bibitem[{Kojima et~al.(2023)Kojima, Gu, Reid, Matsuo, and Iwasawa}]{kojima2023largelanguagemodelszeroshot}
Takeshi Kojima, Shixiang~Shane Gu, Machel Reid, Yutaka Matsuo, and Yusuke Iwasawa. 2023.
\newblock \href {https://arxiv.org/abs/2205.11916} {Large language models are zero-shot reasoners}.
\newblock \emph{Preprint}, arXiv:2205.11916.

\bibitem[{Lanham et~al.(2023)Lanham, Chen, Radhakrishnan, Steiner, Denison, Hernandez, Li, Durmus, Hubinger, Kernion et~al.}]{lanham2023measuring}
Tamera Lanham, Anna Chen, Ansh Radhakrishnan, Benoit Steiner, Carson Denison, Danny Hernandez, Dustin Li, Esin Durmus, Evan Hubinger, Jackson Kernion, and 1 others. 2023.
\newblock Measuring faithfulness in chain-of-thought reasoning.
\newblock \emph{arXiv preprint arXiv:2307.13702}.

\bibitem[{Li et~al.(2025)Li, Bao, Ma, Li, Wang, Men, Zhang, Feng, Liu, and Lin}]{li2025mtr}
Xiaoyuan Li, Keqin Bao, Yubo Ma, Moxin Li, Wenjie Wang, Rui Men, Yichang Zhang, Fuli Feng, Dayiheng Liu, and Junyang Lin. 2025.
\newblock Mtr-bench: A comprehensive benchmark for multi-turn reasoning evaluation.
\newblock \emph{arXiv preprint arXiv:2505.17123}.

\bibitem[{Luo et~al.(2026)Luo, Qiu, Jian, Wang, and Wu}]{luo2026gcot}
Guanran Luo, Wentao Qiu, Zhongquan Jian, Meihong Wang, and Qingqiang Wu. 2026.
\newblock Gcot-decoding: Unlocking deep reasoning paths for universal question answering.
\newblock \emph{arXiv preprint arXiv:2604.06794}.

\bibitem[{Madaan et~al.(2023)Madaan, Tandon, Gupta, Hallinan, Gao, Wiegreffe, Alon, Dziri, Prabhumoye, Yang et~al.}]{madaan2023self}
Aman Madaan, Niket Tandon, Prakhar Gupta, Skyler Hallinan, Luyu Gao, Sarah Wiegreffe, Uri Alon, Nouha Dziri, Shrimai Prabhumoye, Yiming Yang, and 1 others. 2023.
\newblock Self-refine: Iterative refinement with self-feedback.
\newblock \emph{Advances in Neural Information Processing Systems}, 36:46534--46594.

\bibitem[{McKenna et~al.(2023)McKenna, Li, Cheng, Hosseini, Johnson, and Steedman}]{mckenna2023sources}
Nick McKenna, Tianyi Li, Liang Cheng, Mohammad Hosseini, Mark Johnson, and Mark Steedman. 2023.
\newblock Sources of hallucination by large language models on inference tasks.
\newblock In \emph{Findings of the Association for Computational Linguistics: EMNLP 2023}, pages 2758--2774.

\bibitem[{Pearl(2009)}]{pearl2009causality}
Judea Pearl. 2009.
\newblock \emph{Causality: Models, Reasoning, and Inference}.
\newblock Cambridge University Press.

\bibitem[{Radhakrishnan et~al.(2023)Radhakrishnan, Nguyen, Chen, Chen, Denison, Hernandez, Durmus, Hubinger, Kernion, Luko{\v{s}}i{\=u}t{\.e} et~al.}]{radhakrishnan2023question}
Ansh Radhakrishnan, Karina Nguyen, Anna Chen, Carol Chen, Carson Denison, Danny Hernandez, Esin Durmus, Evan Hubinger, Jackson Kernion, Kamil{\.e} Luko{\v{s}}i{\=u}t{\.e}, and 1 others. 2023.
\newblock Question decomposition improves the faithfulness of model-generated reasoning.
\newblock \emph{arXiv preprint arXiv:2307.11768}.

\bibitem[{Sel et~al.(2023)Sel, Al-Tawaha, Khattar, Jia, and Jin}]{sel2023algorithm}
Bilgehan Sel, Ahmad Al-Tawaha, Vanshaj Khattar, Ruoxi Jia, and Ming Jin. 2023.
\newblock Algorithm of thoughts: Enhancing exploration of ideas in large language models.
\newblock \emph{arXiv preprint arXiv:2308.10379}.

\bibitem[{Shen et~al.(2025)Shen, Wang, Tan, Yao, Zhao, Xu, Wang, and Chen}]{shen2025faithcot}
Xu~Shen, Song Wang, Zhen Tan, Laura Yao, Xinyu Zhao, Kaidi Xu, Xin Wang, and Tianlong Chen. 2025.
\newblock Faithcot-bench: Benchmarking instance-level faithfulness of chain-of-thought reasoning.
\newblock \emph{arXiv preprint arXiv:2510.04040}.

\bibitem[{Shen et~al.(2026)Shen, Liu, Shen, Wu, Kong, Huan, and Wang}]{shen2026double}
Yuhao Shen, Tianyu Liu, Junyi Shen, Jinyang Wu, Quan Kong, Li~Huan, and Cong Wang. 2026.
\newblock Double: Breaking the acceleration limit via double retrieval speculative parallelism.
\newblock \emph{arXiv preprint arXiv:2601.05524}.

\bibitem[{Sun et~al.(2025{\natexlab{a}})Sun, Gao, Lin, Ma, and Zhang}]{sun2025explainable}
Yuxi Sun, Wei Gao, Hongzhan Lin, Jing Ma, and Wenxuan Zhang. 2025{\natexlab{a}}.
\newblock Explainable ethical assessment on human behaviors by generating conflicting social norms.
\newblock In \emph{Proceedings of the 14th International Joint Conference on Natural Language Processing and the 4th Conference of the Asia-Pacific Chapter of the Association for Computational Linguistics}, pages 166--184.

\bibitem[{Sun et~al.(2025{\natexlab{b}})Sun, Zuo, Gao, and Ma}]{sun2025causalabstain}
Yuxi Sun, Aoqi Zuo, Wei Gao, and Jing Ma. 2025{\natexlab{b}}.
\newblock Causalabstain: Enhancing multilingual llms with causal reasoning for trustworthy abstention.
\newblock In \emph{Findings of the Association for Computational Linguistics: ACL 2025}, pages 14060--14076.

\bibitem[{Talmor et~al.(2019)Talmor, Herzig, Lourie, and Berant}]{talmor2019commonsenseqa}
Alon Talmor, Jonathan Herzig, Nicholas Lourie, and Jonathan Berant. 2019.
\newblock Commonsenseqa: A question answering challenge targeting commonsense knowledge.
\newblock In \emph{Proceedings of the 2019 Conference of the North American Chapter of the Association for Computational Linguistics: Human Language Technologies, Volume 1 (Long and Short Papers)}, pages 4149--4158.

\bibitem[{Turpin et~al.(2024)Turpin, Michael, Perez, and Bowman}]{turpin2024language}
Miles Turpin, Julian Michael, Ethan Perez, and Samuel~R Bowman. 2024.
\newblock Language models don't always say what they think: Unfaithful explanations in chain-of-thought prompting.
\newblock \emph{Advances in Neural Information Processing Systems (NeurIPS)}.

\bibitem[{Wang et~al.(2025)Wang, Zhang, Wang, Zhao, Feng, He, and Chua}]{wang2025think}
Chengbing Wang, Yang Zhang, Wenjie Wang, Xiaoyan Zhao, Fuli Feng, Xiangnan He, and Tat-Seng Chua. 2025.
\newblock Think-while-generating: On-the-fly reasoning for personalized long-form generation.
\newblock \emph{arXiv preprint arXiv:2512.06690}.

\bibitem[{Wang et~al.(2022)Wang, Wei, Schuurmans, Le, Chi, Narang, Chowdhery, and Zhou}]{wang2022self}
Xuezhi Wang, Jason Wei, Dale Schuurmans, Quoc Le, Ed~Chi, Sharan Narang, Aakanksha Chowdhery, and Denny Zhou. 2022.
\newblock Self-consistency improves chain of thought reasoning in language models.
\newblock \emph{arXiv preprint arXiv:2203.11171}.

\bibitem[{Wang et~al.(2023)Wang, Wei, Schuurmans, Le, Chi, Narang, Chowdhery, and Zhou}]{wang2023selfconsistencyimproveschainthought}
Xuezhi Wang, Jason Wei, Dale Schuurmans, Quoc Le, Ed~Chi, Sharan Narang, Aakanksha Chowdhery, and Denny Zhou. 2023.
\newblock \href {https://arxiv.org/abs/2203.11171} {Self-consistency improves chain of thought reasoning in language models}.
\newblock \emph{Preprint}, arXiv:2203.11171.

\bibitem[{Wang and Huang(2024)}]{wang2024word}
Yu~Wang and Chu-Ren Huang. 2024.
\newblock Word boundary decision: An efficient approach for low-resource word segmentation.
\newblock In \emph{Proceedings of the 38th Pacific Asia Conference on Language, Information and Computation}, pages 160--169.

\bibitem[{Wei et~al.(2022)Wei, Wang, Schuurmans, Bosma, Chi, Le, and Zhou}]{wei2022chain}
Jason Wei, Xuezhi Wang, Dale Schuurmans, Maarten Bosma, Ed~Chi, Quoc Le, and Denny Zhou. 2022.
\newblock Chain-of-thought prompting elicits reasoning in large language models.
\newblock In \emph{Advances in Neural Information Processing Systems (NeurIPS)}.

\bibitem[{Wei et~al.(2023)Wei, Wang, Schuurmans, Bosma, Ichter, Xia, Chi, Le, and Zhou}]{wei2023chainofthoughtpromptingelicitsreasoning}
Jason Wei, Xuezhi Wang, Dale Schuurmans, Maarten Bosma, Brian Ichter, Fei Xia, Ed~Chi, Quoc Le, and Denny Zhou. 2023.
\newblock \href {https://arxiv.org/abs/2201.11903} {Chain-of-thought prompting elicits reasoning in large language models}.
\newblock \emph{Preprint}, arXiv:2201.11903.

\bibitem[{Wu et~al.(2026)Wu, Yang, Yang, Shen, Zhang, Wen, and Tao}]{wu2026spark}
Jinyang Wu, Shuo Yang, Changpeng Yang, Yuhao Shen, Shuai Zhang, Zhengqi Wen, and Jianhua Tao. 2026.
\newblock Spark: Strategic policy-aware exploration via dynamic branching for long-horizon agentic learning.
\newblock \emph{arXiv preprint arXiv:2601.20209}.

\bibitem[{Wu et~al.(2024)Wu, Yu, Chen, Wang, Rossi, Kim, Rao, and McAuley}]{wu-etal-2024-decot}
Junda Wu, Tong Yu, Xiang Chen, Haoliang Wang, Ryan Rossi, Sungchul Kim, Anup Rao, and Julian McAuley. 2024.
\newblock \href {https://doi.org/10.18653/v1/2024.acl-long.758} {{D}e{C}o{T}: Debiasing chain-of-thought for knowledge-intensive tasks in large language models via causal intervention}.
\newblock In \emph{Proceedings of the 62nd Annual Meeting of the Association for Computational Linguistics (Volume 1: Long Papers)}, pages 14073--14087, Bangkok, Thailand. Association for Computational Linguistics.

\bibitem[{Xi et~al.(2024)Xi, Jin, Zhou, Zheng, Gao, Gui, Zhang, and Huang}]{xi2024selfpolishenhancereasoninglarge}
Zhiheng Xi, Senjie Jin, Yuhao Zhou, Rui Zheng, Songyang Gao, Tao Gui, Qi~Zhang, and Xuanjing Huang. 2024.
\newblock \href {https://arxiv.org/abs/2305.14497} {Self-polish: Enhance reasoning in large language models via problem refinement}.
\newblock \emph{Preprint}, arXiv:2305.14497.

\bibitem[{Xi et~al.(2023)Xi, Jin, Zhou, Zheng, Gao, Liu, Gui, Zhang, and Huang}]{xi2023self}
Zhiheng Xi, Senjie Jin, Yuhao Zhou, Rui Zheng, Songyang Gao, Jia Liu, Tao Gui, Qi~Zhang, and Xuan-Jing Huang. 2023.
\newblock Self-polish: Enhance reasoning in large language models via problem refinement.
\newblock In \emph{Findings of the Association for Computational Linguistics: EMNLP 2023}, pages 11383--11406.

\bibitem[{Xiong et~al.(2025)Xiong, Chen, Qi, and Lakkaraju}]{xiong2025measuring}
Zidi Xiong, Shan Chen, Zhenting Qi, and Himabindu Lakkaraju. 2025.
\newblock Measuring the faithfulness of thinking drafts in large reasoning models.
\newblock \emph{arXiv preprint arXiv:2505.13774}.

\bibitem[{Xu et~al.(2026{\natexlab{a}})Xu, Wang, Jia, Wu, Liu, and Dong}]{xu2026rcbsfmultiagentframeworkautomated}
Shijia Xu, Yu~Wang, Xiaolong Jia, Zhou Wu, Kai Liu, and April~Xiaowen Dong. 2026{\natexlab{a}}.
\newblock \href {https://arxiv.org/abs/2604.10740} {Rcbsf: A multi-agent framework for automated contract revision via stackelberg game}.
\newblock \emph{Preprint}, arXiv:2604.10740.

\bibitem[{Xu et~al.(2026{\natexlab{b}})Xu, Wu, Jia, Wang, Liu, and Dong}]{xu2026selfcorrectingragenhancingfaithfulness}
Shijia Xu, Zhou Wu, Xiaolong Jia, Yu~Wang, Kai Liu, and April~Xiaowen Dong. 2026{\natexlab{b}}.
\newblock \href {https://arxiv.org/abs/2604.10734} {Self-correcting rag: Enhancing faithfulness via mmkp context selection and nli-guided mcts}.
\newblock \emph{Preprint}, arXiv:2604.10734.

\bibitem[{Yang et~al.(2025)Yang, Lee, Kassner, Gottesman, Riedel, and Geva}]{yang2025well}
Sohee Yang, Sang-Woo Lee, Nora Kassner, Daniela Gottesman, Sebastian Riedel, and Mor Geva. 2025.
\newblock How well can reasoning models identify and recover from unhelpful thoughts?
\newblock \emph{arXiv preprint arXiv:2506.10979}.

\bibitem[{Yao et~al.(2023)Yao, Yu, Zhao, Shafran, Griffiths, Cao, and Narasimhan}]{yao2023tree}
Shunyu Yao, Dian Yu, Jeffrey Zhao, Izhak Shafran, Tom Griffiths, Yuan Cao, and Karthik Narasimhan. 2023.
\newblock Tree of thoughts: Deliberate problem solving with large language models.
\newblock \emph{Advances in neural information processing systems}, 36:11809--11822.

\bibitem[{Yu et~al.(2023)Yu, He, and Ying}]{yu2023thought}
Junchi Yu, Ran He, and Rex Ying. 2023.
\newblock Thought propagation: An analogical approach to complex reasoning with large language models.
\newblock \emph{arXiv preprint arXiv:2310.03965}.

\bibitem[{Yu et~al.(2025)Yu, Wang, Yang, Li, Liu, Xue, Wang, and Yang}]{yu2025causalsufficiencynecessityimproves}
Xiangning Yu, Zhuohan Wang, Linyi Yang, Haoxuan Li, Anjie Liu, Xiao Xue, Jun Wang, and Mengyue Yang. 2025.
\newblock \href {https://arxiv.org/abs/2506.09853} {Causal sufficiency and necessity improves chain-of-thought reasoning}.
\newblock \emph{Preprint}, arXiv:2506.09853.

\bibitem[{Zhang et~al.(2024)Zhang, Zhang, and Zhou}]{zhang2024causal}
Chen Zhang, Lanning Zhang, and Dexiang Zhou. 2024.
\newblock Causal walk: Debiasing multi-hop fact verification with front-door adjustment.
\newblock In \emph{Proceedings of the AAAI Conference on Artificial Intelligence}, volume~38, pages 19533--19541.

\bibitem[{Zhang et~al.(2023)Zhang, Zhang, Hou, Fan, Li, Liu, Zhang, and Chang}]{zhang2023certified}
Zhen Zhang, Guanhua Zhang, Bairu Hou, Wenqi Fan, Qing Li, Sijia Liu, Yang Zhang, and Shiyu Chang. 2023.
\newblock Certified robustness for large language models with self-denoising.
\newblock \emph{arXiv preprint arXiv:2307.07171}.

\bibitem[{Zheng et~al.(2023)Zheng, Chiang, Sheng, Zhuang, Wu, Zhuang, Lin, Li, Li, Xing et~al.}]{zheng2023judging}
Lianmin Zheng, Wei-Lin Chiang, Ying Sheng, Siyuan Zhuang, Zhanghao Wu, Yonghao Zhuang, Zi~Lin, Zhuohan Li, Dacheng Li, Eric Xing, and 1 others. 2023.
\newblock Judging llm-as-a-judge with mt-bench and chatbot arena.
\newblock \emph{Advances in neural information processing systems}, 36:46595--46623.

\bibitem[{Zhou et~al.(2022)Zhou, Sch{\"a}rli, Hou, Wei, Scales, Wang, Schuurmans, Cui, Bousquet, Le et~al.}]{zhou2022least}
Denny Zhou, Nathanael Sch{\"a}rli, Le~Hou, Jason Wei, Nathan Scales, Xuezhi Wang, Dale Schuurmans, Claire Cui, Olivier Bousquet, Quoc Le, and 1 others. 2022.
\newblock Least-to-most prompting enables complex reasoning in large language models.
\newblock \emph{arXiv preprint arXiv:2205.10625}.

\bibitem[{Zhou et~al.(2024)Zhou, Tao, Zhu, Luo, Wang, and Han}]{zhou2024can}
Zhanke Zhou, Rong Tao, Jianing Zhu, Yiwen Luo, Zengmao Wang, and Bo~Han. 2024.
\newblock Can language models perform robust reasoning in chain-of-thought prompting with noisy rationales?
\newblock \emph{Advances in Neural Information Processing Systems}, 37:123846--123910.

\bibitem[{Zhu et~al.(2023)Zhu, Thomason, and Jia}]{zhu-etal-2023-chain}
Wang Zhu, Jesse Thomason, and Robin Jia. 2023.
\newblock \href {https://doi.org/10.18653/v1/2023.emnlp-main.547} {Chain-of-questions training with latent answers for robust multistep question answering}.
\newblock In \emph{Proceedings of the 2023 Conference on Empirical Methods in Natural Language Processing}, pages 8845--8860, Singapore. Association for Computational Linguistics.

\end{thebibliography}

\appendix
\appendix
\newpage
\section*{Appendix}

\section{More Details of FACT-E} 

\subsection{The Standard and Lightweight Versions}

Algorithm~\ref{algorithm2} presents the standard version of FACT-E, which evaluates all sequential dependencies between reasoning steps. To reduce inference cost, we also introduce a lightweight variant (Algorithm~\ref{alg:cot_causal_rerank_light}) based on a fixed-checkpoint mechanism. In the lightweight version, the number of inspected checkpoints is set to $N$, matching the sampling budget used in the consistency estimation step. The specific logical perturbations used for causal verification are described in Appendix~\ref{noise_inject}.

\subsection{Algorithm for Detecting Noisy CoT Methods}

As detailed in Algorithm~\ref{alg:denoise}, FACT-E identifies and filters unreliable rationales from a benchmark-provided noisy prompt set $\mathcal{P}_{\mathrm{noise}}$ associated with the current test query $q_{\mathrm{test}}$. The size of $\mathcal{P}_{\mathrm{noise}}$ may vary across test instances, reflecting different noisy-prompt settings in the NoRa benchmarks. The procedure consists of two phases: a denoising stage that constructs a refined exemplar set $\mathcal{P}_{\mathrm{clean}}$, followed by a final inference stage.

Phase 1 aims to detect noisy demonstrations.
For each exemplar $(q_i,\mathbf{S}^{(i)},a_i)\in\mathcal{P}_{\mathrm{noise}}$, FACT-E performs a three-step assessment.
(1) \textit{CoT-to-Answer Consistency} ($\mathcal{C}_{\mathbf{S}}$): We evaluate whether the provided rationale is externally aligned with its associated label by computing
$\mathcal{C}_{\mathbf{S}^{(i)}} \gets P(a_i \mid q_i,\mathbf{S}^{(i)}).$
Exemplars with zero consistency, i.e., $\mathcal{C}_{\mathbf{S}^{(i)}}=0$, are pruned immediately.
(2) \textit{Intra-Chain Faithfulness} ($\mathcal{F}_{\mathbf{S}}$): To efficiently assess step-level causal validity, we sample a subset of split points
$
\mathcal{T}^{(i)} \subseteq \{1,\ldots,L_i-1\}, \qquad |\mathcal{T}^{(i)}|=\min(N,L_i-1),
$
where $L_i=|\mathbf{S}^{(i)}|$. For each sampled step $t\in\mathcal{T}^{(i)}$, we generate perturbed continuations $\mathbf{S}_{>t}^{\prime\,(e_j)}$ via noise injection and estimate faithfulness by checking whether the model prefers the original continuation over its perturbed counterparts under the same prefix.
(3) \textit{Reliability filtering via FACT-E}: We compute the final reliability score
$
\mathcal{R}_{\mathbf{S}^{(i)}} \gets \mathcal{C}_{\mathbf{S}^{(i)}} \cdot \mathcal{F}_{\mathbf{S}^{(i)}}.
$
Only exemplars satisfying $\mathcal{R}_{\mathbf{S}^{(i)}} \ge \tau$ are retained in $\mathcal{P}_{\mathrm{clean}}$. In practice, the threshold $\tau$ is selected on validation data. Phase 2 is the final inference.
The LLM then performs few-shot inference on the test query $q_{\mathrm{test}}$ using the filtered prompt set $\mathcal{P}_{\mathrm{clean}}$, ensuring that the prediction is conditioned only on more reliable demonstrations.

\begin{algorithm}[t!]
\footnotesize
\caption{Select the Trustworthy CoT via FACT-E (Standard)}
\label{algorithm2}
\begin{algorithmic}[1]
\Require 
    Candidate set $\mathcal{S}=\{(\mathbf{S}^{(j)},a^{(j)})\}_{j=1}^{K}$;
    query $q$;
    language model $\mathcal{M}$;
    perturbation set $\mathcal{E}$;
    sampling budget $N$
\Ensure 
    Optimal CoT $\mathbf{S}_{\mathrm{opt}}$,
    answer $a_{\mathrm{opt}}$,
    score $\mathcal{R}_{\max}$

\State $\mathrm{SC} \gets \emptyset$

\For{each candidate $(\mathbf{S}^{(j)},a^{(j)}) \in \mathcal{S}$}

    \State \textcolor{purple}{// \textbf{Step 1}: CoT-to-Answer Consistency}
    \State $\mathcal{C}_{\mathbf{S}^{(j)}} \gets \frac{1}{N}\sum_{n=1}^{N} P(J^{(n)}=1 \mid q,\mathbf{S}^{(j)}) \cdot \mathbf{1}\{J^{(n)}=1\}$
    \If{$\mathcal{C}_{\mathbf{S}^{(j)}} = 0$}
        \State \textbf{continue}
    \EndIf

    \State \textcolor{purple}{// \textbf{Step 2}: Intra-Chain Faithfulness}
    \State $L_j \gets |\mathbf{S}^{(j)}|$
    \For{$t = 1$ \textbf{to} $L_j-1$}
        \State $\mathbf{S}_{>t}^{\prime\,(e)} \gets \mathcal{M}(\cdot \mid \mathrm{InjectNoise}(\mathbf{S}_{\le t}^{(j)}, e)), \ \forall e \in \mathcal{E}$
        \State $P_{\mathrm{faith}}^{(j,t)} \gets \mathbb{E}_{e \in \mathcal{E}} \left[
        \mathbf{1}\!\left[
        (\mathbf{S}_{\le t}^{(j)},\mathbf{S}_{>t}^{(j)})
        \succ
        (\mathbf{S}_{\le t}^{(j)},\mathbf{S}_{>t}^{\prime\,(e)})
        \right]\right]$
    \EndFor
    \State $\mathcal{F}_{\mathbf{S}^{(j)}} \gets \frac{1}{L_j-1}\sum_{t=1}^{L_j-1} P_{\mathrm{faith}}^{(j,t)}$

    \State \textcolor{purple}{// \textbf{Step 3}: FACT-E score}
    \State $\mathcal{R}_{\mathbf{S}^{(j)}} \gets \mathcal{C}_{\mathbf{S}^{(j)}} \cdot \mathcal{F}_{\mathbf{S}^{(j)}}$
    \State $\mathrm{SC} \gets \mathrm{SC} \cup \{(\mathbf{S}^{(j)},a^{(j)},\mathcal{R}_{\mathbf{S}^{(j)}})\}$
\EndFor

\State $(\mathbf{S}_{\mathrm{opt}},a_{\mathrm{opt}},\mathcal{R}_{\max}) \gets \arg\max_{(\mathbf{S},a,\mathcal{R}_{\mathbf{S}})\in \mathrm{SC}} \mathcal{R}_{\mathbf{S}}$
\State \Return $\mathbf{S}_{\mathrm{opt}},a_{\mathrm{opt}},\mathcal{R}_{\max}$
\end{algorithmic}
\end{algorithm}

\begin{algorithm}[t!]
\footnotesize
\caption{Detecting and Filtering Noisy CoT to Enhance Answering (RQ3)}
\label{alg:denoise}
\begin{algorithmic}[1]
\Require
    Noisy prompt set $\mathcal{P}_{\mathrm{noise}}=\{(q_i,\mathbf{S}^{(i)},a_i)\}_{i=1}^{|\mathcal{P}_{\mathrm{noise}}|}$;
    Test query $q_{\mathrm{test}}$;
    Language model $\mathcal{M}$;
    Perturbation set $\mathcal{E}$;
    Sampling budget $N$;
    Threshold $\tau$
\Ensure
    Final prediction $a_{\mathrm{test}}$ based on the filtered prompt set $\mathcal{P}_{\mathrm{clean}}$

\State $\mathcal{P}_{\mathrm{clean}} \gets \emptyset$

\State \textcolor{purple}{// \textbf{Phase 1}: Detect noisy demonstrations}
\For{each exemplar $(q_i,\mathbf{S}^{(i)},a_i)\in\mathcal{P}_{\mathrm{noise}}$}

    \State \textcolor{purple}{// \textbf{Step 1}: CoT-to-Answer Consistency}
    \State $\mathcal{C}_{\mathbf{S}^{(i)}} \gets P(a_i \mid q_i,\mathbf{S}^{(i)})$
    \If{$\mathcal{C}_{\mathbf{S}^{(i)}} = 0$}
        \State \textbf{continue}
    \EndIf

    \State \textcolor{purple}{// \textbf{Step 2}: Intra-Chain Faithfulness}
    \State $L_i \gets |\mathbf{S}^{(i)}|$
    \State Sample $\mathcal{T}^{(i)} \subseteq \{1,\ldots,L_i-1\}$ such that $|\mathcal{T}^{(i)}|=\min(N,L_i-1)$
    \For{each $t \in \mathcal{T}^{(i)}$}
        \State $\mathbf{S}_{>t}^{\prime\,(e_j)} \gets \mathcal{M}(\cdot \mid \mathrm{InjectNoise}(\mathbf{S}_{\le t}^{(i)}, e_j)), \ \forall e_j \in \mathcal{E}$
        \State $P_{\mathrm{faith}}^{(i,t)} \gets \frac{1}{|\mathcal{E}|}\sum_{e_j\in\mathcal{E}}
        \mathbf{1}\!\left[
        (\mathbf{S}_{\le t}^{(i)},\mathbf{S}_{>t}^{(i)})
        \succ
        (\mathbf{S}_{\le t}^{(i)},\mathbf{S}_{>t}^{\prime\,(e_j)})
        \right]$
    \EndFor
    \State $\mathcal{F}_{\mathbf{S}^{(i)}} \gets \frac{1}{|\mathcal{T}^{(i)}|}\sum_{t\in\mathcal{T}^{(i)}} P_{\mathrm{faith}}^{(i,t)}$

    \State \textcolor{purple}{// \textbf{Step 3}: Reliability filtering}
    \State $\mathcal{R}_{\mathbf{S}^{(i)}} \gets \mathcal{C}_{\mathbf{S}^{(i)}} \cdot \mathcal{F}_{\mathbf{S}^{(i)}}$
    \If{$\mathcal{R}_{\mathbf{S}^{(i)}} \ge \tau$}
        \State $\mathcal{P}_{\mathrm{clean}} \gets \mathcal{P}_{\mathrm{clean}} \cup \{(q_i,\mathbf{S}^{(i)},a_i)\}$
    \EndIf
\EndFor

\State \textcolor{purple}{// \textbf{Phase 2}: Final inference with filtered demonstrations}
\State $a_{\mathrm{test}} \gets \mathcal{M}(\mathcal{P}_{\mathrm{clean}}, q_{\mathrm{test}})$
\State \Return $a_{\mathrm{test}}$
\end{algorithmic}
\end{algorithm}

\subsection{Noise Injection}\label{noise_inject}
To rigorously evaluate the faithfulness of reasoning, we apply a set of perturbations $\mathcal{E}$ to the CoT candidates, as illustrated in Table~\ref{tab: noise types}. Specifically, Operation Error and Conceptual Swap target the precision of individual steps by altering operators and substituting concepts, respectively. Misgeneralization and Reordered Logic perturb the inductive and structural integrity of the reasoning path. Finally, Contradiction assesses the model's ability to maintain logical grounding by introducing premise-violating information. The detailed prompts of noise injected (construct the counterfactual segments of the chain) are shown in Table~\ref{tab:prompts_our}.

\subsection{More Case Studies}

Table 8 illustrates additional case studies about the score of FACT-E regarding CoT. CoTs with logical errors (highlighted in red) and incorrect answers receive significantly low scores (e.g., 0.177 and 0.338), ensuring that hallucinatory or flawed reasoning is penalized. Correct answers derived from rigorous, error-free reasoning steps achieve the highest scores (approx. 0.8), validating the metric's ability to select optimal CoTs. Crucially, the FACT-E distinguishes between "correct answer with flawed logic" and "correct answer with sound logic." In the third case, while both paths yield the correct result, the mathematically rigorous chain scores higher (0.7995) than the one containing minor logical defects (0.5992). Similarly, in the third coordinate conversion task, FACT-E identifies that the deduction following "so" does not maintain a strict causal relationship with the preceding steps, assigning a score of 0.5992 compared to the more logically sound CoT 2, which scores 0.7995. 

\subsection{LLM Inference Overhead}

For efficiency analysis, Table~\ref{table:overhead} reports the number of LLM inference requests per query under the experimental setting with sampling budget $N=3$ and a single perturbation type per inspected split point. Under this setting, the lightweight variant scales linearly with the checkpoint budget $N$, whereas the standard variant scales linearly with the number of inspected split points in the reasoning chain. In practice, the exact number of requests may still vary depending on whether multiple operations are batched into a single prompt.

\begin{table}[t!]
    \small
    \centering
    \begin{tabular}{l|c}
        \toprule
        Method & \# LLM inference requests \\
        \hline
        \textsc{Polish} & 7 \\
        \textsc{Consistency} & 3 \\
        \textsc{Reflect} & 6 \\
        \textsc{Denoise} & 7 \\
        \hline
        \textsc{FACT-E}$_{\text{lightweight}}$ & 7 \\
        \textsc{FACT-E}$_{\text{standard}}$ & $3 \cdot \ell(c) + 1$ \\
        \bottomrule
    \end{tabular}
    \caption{Number of LLM inference requests per query under $N=3$ and a single perturbation type per inspected split point. Here $\ell(c)$ denotes the number of reasoning steps in the CoT. The exact count may vary depending on prompt batching in implementation.}
    \label{table:overhead}
\end{table}

\begin{table*}[t!]
\centering
\small
\label{tab:unified_noise_definitions}
\begin{tabular}{lp{5.5cm}p{5.5cm}}
\toprule
\textbf{Noise Type} & \textbf{Description (MATH Task)} & \textbf{Description (Commonsense)} \\ \midrule
Operation Error     & Modification of a specific mathematical operation or procedural step. & Modification of a specific logical step or operative element within the reasoning. \\ \addlinespace
Conceptual Swap     & Substitution of distinct mathematical or logical concepts. & Substitution of semantically or logically related entities, properties, or concepts. \\ \addlinespace
Misgeneralization   & Erroneous extrapolation from a specific concept to an invalid general rule. & Improper extension of a specific concept or heuristic to an invalid or broader context. \\ \addlinespace
Reordered Logic     & Permutation of the sequential order of reasoning steps. & Permutation of the sequential or causal order of reasoning steps. \\ \addlinespace
Contradiction       & Introduction of an inconsistency with established premises or prior conclusions. & Introduction of an assertion that conflicts with established facts, premises, or prior logic. \\ \bottomrule
\end{tabular}
\caption{Unified Definitions of Noise Types across MATH and Commonsense Tasks.}
\label{tab: noise types}
\end{table*}

\subsection{Ablative Study}
\begin{figure}[t!]
    \centering
    \includegraphics[width=1\linewidth]{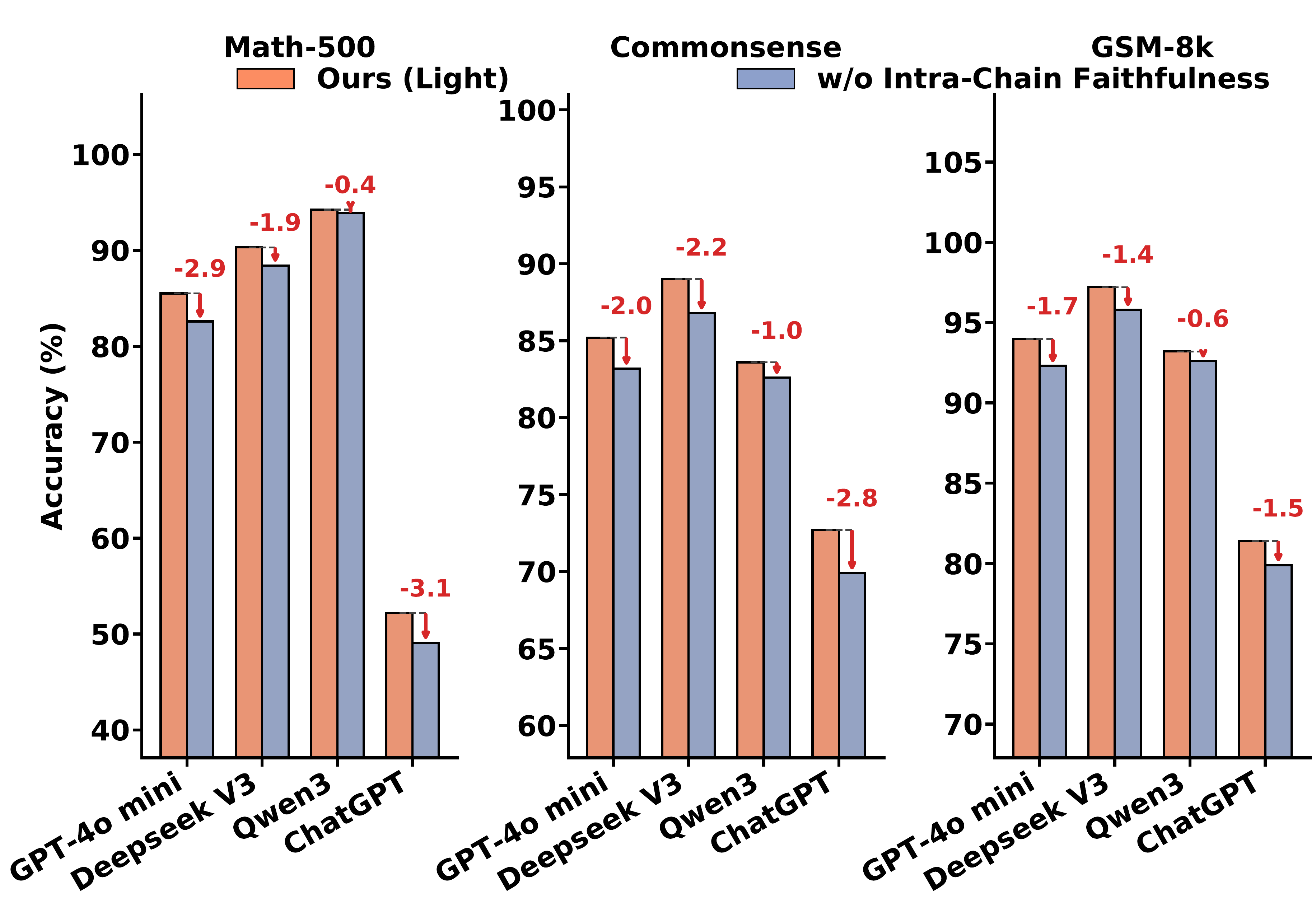}
    \caption{Ablation study of FACT-E.}
    \label{fig:ablative}
\end{figure}

The ablation results in Figure~\ref{fig:ablative} further confirm the effectiveness of combining Intra-Chain Faithfulness and CoT-to-Answer Consistency. 

\subsection{Baseline Methods}\label{appendix: baselines}
We conduct evaluations using four LLM backbones, including DeepSeek-V3, Qwen3-14B, Gpt-4o-mini, and Gpt-3.5-turbo. For all models, we set the temperature parameter $\tau$ to 0. To ensure stable results, we evaluate 500 questions per task and repeat each experiment three times. We compare against five representative baseline methods. \textsc{Polish}~\citep{xi2023self} and \textsc{Reflect}~\citep{kadavath2022language} fall under the self-correction paradigm, aiming to improve generation quality through prompt rephrasing and iterative reflection. \textsc{Denoise}~\citep{zhang2023certified} adopts a mask-reconstruction strategy that requires the model to recover masked content, while \textsc{Consistency}~\citep{wang2022self} aggregates multiple sampled outputs to improve robustness. 

\begin{itemize}
\item Self-Polish (\textsc{Polish})~\citep{xi2023self} enhances the quality of reasoning chains by teaching large language models (LLMs) to eliminate noisy information, restructure logical sequences, and reorganize local conditions into coherent reasoning steps. In our implementation, we (1) prompt the LLM to independently refine each noisy chain-of-thought (CoT) exemplar without external guidance, repeating this refinement process three times, and (2) aggregate these rephrased demonstrations to construct the final context for downstream task reasoning.

\item Self-Denoise (\textsc{Denoise})~\citep{zhang2023certified} improves LLM robustness by preprocessing input prompts with random masking, requiring the model to reconstruct the masked content. This process reduces noise and mitigates incoherent reasoning. Our implementation involves (1) applying random masks to noisy rationales in each demonstration, (2) prompting the LLM to infer and complete the masked sections for each example, and (3) utilizing the reconstructed CoT demonstrations for subsequent task reasoning. This procedure is repeated three times, with the most frequent answer selected as final.

\item Self-Consistency (\textsc{Consistency})~\citep{wang2022self} enhances reasoning performance through output sampling and majority voting, without modifying the input. In our approach, we execute the same reasoning task three times and select the answer that appears most frequently across all runs.

\item Self-Reflect (\textsc{Reflect})~\citep{kadavath2022language} enhances LLM reasoning by encouraging the model to explicitly critique and revise its own intermediate outputs. In our implementation, we (1) prompt the LLM to generate an initial CoT, (2) instruct the model to self-reflect on potential flaws or logical gaps, and (3) direct the LLM to output a final revised CoT. 

\end{itemize}

\begin{table*}[t!]
\centering
\scriptsize 
\begin{tabular}{p{0.1cm}p{0.8cm}p{13cm}p{0.2cm}}
\toprule
\textbf{\#} & \textbf{Answer} & \textbf{CoT (Chain of Thought)} & \textbf{Score} \\
\midrule

\rowcolor{casestudy}
\multicolumn{4}{p{\dimexpr\textwidth-2\tabcolsep\relax}}{%
Question: Suppose $\sin D = 0.7$ in the diagram below. What is $DE$? [asy] pair D,E,F; F = (0,0); D = (sqrt(51),7); E = (0,7); draw(D--E--F--D); draw(rightanglemark(D,E,F,15)); label("$D$",D,NE); label("$E$",E,NW); label("$F$",F,SW); label("$7$",(E+F)/2,W); [/asy]
} \\

1 & $6.86$ \textcolor{red}{\XSolidBrush} &
We are given that $\sin D = 0.7$ and we need to find the length of $DE$. In the right triangle $DEF$, \textcolor{wrong}{we know that $DE$ is the side opposite angle $D$ and $EF$ is the side adjacent to angle $D$}. The length of $EF$ is given as $7$. Using the definition of sine, we have: $\sin D = \frac{\text{opposite}}{\text{hypotenuse}} = \frac{DE}{DF}$. We can also express $DF$ in terms of $DE$ and $EF$ using the Pythagorean theorem: $DF = \sqrt{DE^2 + EF^2}$. Substituting $EF = 7$ into the equation gives: $DF = \sqrt{DE^2 + 7^2} = \sqrt{DE^2 + 49}$. Now substituting this into the sine equation: $0.7 = \frac{DE}{\sqrt{DE^2 + 49}}$. To eliminate the square root, we can square both sides: $0.7^2 = \frac{DE^2}{DE^2 + 49}$. Calculating $0.7^2$: $0.49 = \frac{DE^2}{DE^2 + 49}$. Cross-multiplying gives: $0.49(DE^2 + 49) = DE^2$. Expanding this: $0.49DE^2 + 24.01 = DE^2$. Rearranging terms: $DE^2 - 0.49DE^2 = 24.01$. This simplifies to: $0.51DE^2 = 24.01$. Dividing both sides by $0.51$: $DE^2 = \frac{24.01}{0.51} \approx 47.0$. Taking the square root to find $DE$: $DE \approx \sqrt{47.0} \approx 6.86$.
& 0.338 \\

\midrule

\rowcolor{casestudy}
\multicolumn{4}{p{\dimexpr\textwidth-2\tabcolsep\relax}}{%
Question: $\overline{BC}$ is parallel to the segment through $A$, and $AB = BC$. What is the number of degrees represented by $x$? [asy] draw((0,0)--(10,0)); draw((0,3)--(10,3)); draw((2,3)--(8,0)); draw((2,3)--(4,0)); label("$A$",(2,3),N); label("$B$",(4,0),S); label("$C$",(8,0),S); label("$124^{\circ}$",(2,3),SW); label("$x^{\circ}$",(4.5,3),S); [/asy]} \\

1 & $124^\circ$ \textcolor{red}{\XSolidBrush} &
We start by analyzing the given information. We have two parallel lines, one at the top (where point A is located) and one at the bottom (where points B and C are located). The angle at point A is given as $124^\circ$. \textcolor{wrong}{Since $\overline{BC}$ is parallel to the line through $A$, the angle at $A$ and the angle $x$ at point $B$ are corresponding angles. According to the properties of parallel lines, corresponding angles are equal.} Therefore, we can set up the equation: $x = 124^\circ$. Thus, the value of $x$ is directly equal to the angle at point $A$.
& 0.398 \\

2 & $28^\circ$ \textcolor{green}{\Checkmark} &
In the given problem, we have two parallel lines represented by the segments through points $A$ and $B$ and $C$. Since $\overline{BC}$ is parallel to the segment through $A$, we can use the properties of alternate interior angles. The angle at point $A$ is given as $124^\circ$. Since $AB$ is equal to $BC$, triangle $ABC$ is isosceles with $AB = BC$. Therefore, the angles at points $B$ and $C$ must be equal. Let the angle at point $B$ be $x$. The angle at point $C$ will also be $x$. The sum of the angles in triangle $ABC$ must equal $180^\circ$. Thus, we can set up the equation: $124^\circ + x + x = 180^\circ$ This simplifies to: $124^\circ + 2x = 180^\circ$ Subtracting $124^\circ$ from both sides gives: $2x = 180^\circ - 124^\circ = 56^\circ$ Dividing both sides by $2$ yields: $x = \frac{56^\circ}{2} = 28^\circ$. Thus, the value of $x$ is $28^\circ$.
& 0.798 \\

3 & $56^\circ$ \textcolor{red}{\XSolidBrush} &
In the given problem, we have two parallel lines, and we know that $\overline{BC}$ is parallel to the segment through $A$. The angle at point $A$ is given as $124^\circ$. Since $AB$ is equal to $BC$, triangle $ABC$ is isosceles with $AB = BC$. \textcolor{wrong}{The angle $x$ is the angle at point $B$ which is opposite to the angle at point $A$. By the properties of parallel lines and the angles formed, we can use the fact that the angle at $A$ and the angle at $B$ are supplementary because they are on the same side of the transversal line $AC$.} Thus, we can set up the equation: $x + 124^\circ = 180^\circ$ Now, we can solve for $x$: $x = 180^\circ - 124^\circ = 56^\circ$ Therefore, the value of $x$ is $56^\circ$.
& 0.177 \\

\midrule

\rowcolor{casestudy}
\multicolumn{4}{p{\textwidth}}{%
Question: Convert the point $(0,3)$ in rectangular coordinates to polar coordinates. Enter your answer in the form $(r,\theta),$ where $r > 0$ and $0 \le \theta < 2 \pi.$} \\

1 & $(3, \frac{\pi}{2})$ \textcolor{green}{\Checkmark} &
To convert from rectangular coordinates $(x,y)$ to polar coordinates $(r,\theta),$ we use the formulas $r = \sqrt{x^2 + y^2}$ and $\theta = \arctan\left(\frac{y}{x}\right).$ Given the point $(0,3),$ we have $r = \sqrt{0^2 + 3^2} = 3$ and $\theta = \arctan\left(\frac{3}{0}\right) = \frac{\pi}{2}.$ \textcolor{wrong}{However, we need to adjust $\theta$ to be in the interval $[0, 2\pi),$ so the final answer is $(3, \frac{\pi}{2}).$}
& 0.5992 \\

2 & $(3, \frac{\pi}{2})$ \textcolor{green}{\Checkmark} &
To convert a point from rectangular coordinates $(x,y)$ to polar coordinates $(r,\theta)$, we use the formulas $r = \sqrt{x^2 + y^2}$ and $\theta = \arctan\left(\frac{y}{x}\right)$. Given the point $(0,3)$, we have $x = 0$ and $y = 3$. Calculating $r = \sqrt{0^2 + 3^2} = 3$ and $\theta = \arctan\left(\frac{3}{0}\right) = \frac{\pi}{2}$ (since $\arctan$ is undefined at $x=0$ but we know the point lies on the positive $y$-axis). Therefore, the point $(0,3)$ in rectangular coordinates is equivalent to the point $(3,\frac{\pi}{2})$ in polar coordinates.
& 0.7995 \\

\bottomrule
\end{tabular}
\caption{Question and Answer Evaluation Results, the flawed part of CoT is highlighted in \textcolor{wrong}{red}. The correct and wrong answers are shown in \textcolor{green}{\Checkmark} and \textcolor{red}{\XSolidBrush}, respectively.}
\label{app:cases}
\end{table*}

\section{Further Analysis}
\begin{figure}[t!]
    \centering
    \includegraphics[width=1\linewidth]{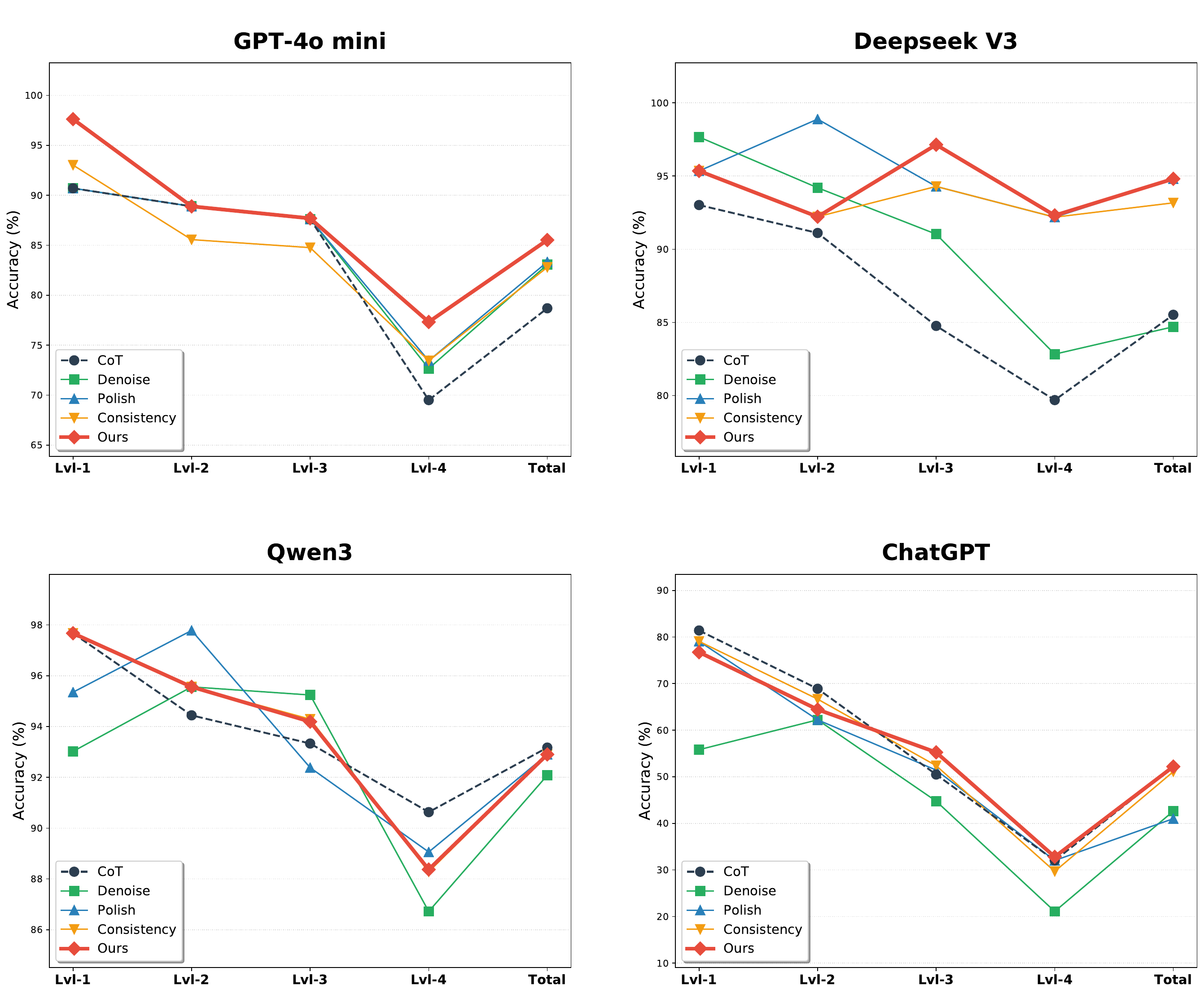}
    \caption{Analysis of different levels of MATH-500.}
    \label{fig:mathlevel}
\end{figure}

As illustrated in the Figure~\ref{fig:mathlevel}, performance universally degrades across all base models as difficulty increases from Lvl-1 to Lvl-4. Notably, $\textsc{Ours}$ exhibits superior robustness, effectively mitigating the "performance cliff" observed in other baselines. On DeepSeek-V3, $\textsc{Ours}$ achieves an accuracy of 92.31\% at the highest difficulty (Lvl-4), outperforming the $\textsc{CoT}$ baseline (79.69\%) by a substantial margin of 12.62\%. While standard enhancement methods like $\textsc{Polish}$ or $\textsc{Consistency}$ yield gains on simpler tasks, their efficacy diminishes as logical complexity peaks. $\textsc{Ours}$ consistently maintains a flatter decay curve by inserting mechanisms, e.g., error-correction/verification. This is particularly evident in the ChatGPT experiments, where $\textsc{Ours}$ preserves its performance edge even when other methods drop below the 33\% accuracy threshold at Lvl-4. Across the full Math500 dataset, $\textsc{Ours}$ achieves state-of-the-art or competitive results on all evaluated LLMs. Compared to the sampling-heavy $\textsc{Consistency}$ baseline, $\textsc{Ours}$ improves total accuracy by 9.52\% on GPT-4o-mini and 4.03\% on Deepseek V3, demonstrating superior reasoning efficiency and reliability across varying model scales.
\begin{table*}[t!]
\centering
\footnotesize
\setlength{\tabcolsep}{3pt}
\begin{tabular}{lcccccc}
\toprule
\cellcolor{white}{} & \multicolumn{2}{c}{5 Examples} & \multicolumn{2}{c}{10 Examples} & \multicolumn{2}{c}{15 Examples} \\
\cmidrule(lr){2-3} \cmidrule(lr){4-5} \cmidrule(lr){6-7}
Method & DeepSeek-V3 & Qwen3-14B & DeepSeek-V3 & Qwen3-14B & DeepSeek-V3 & Qwen3-14B \\
\midrule
Standard CoT & 92.35 & 90.16 & 92.35 & 91.53 & 92.62 & 92.62 \\
Denoise      & 92.62 & 85.79 & 94.54 & 89.34 & 92.08 & 90.16 \\
Polish       & 94.81 & 92.08 & 92.62 & 89.07 & 91.80 & 90.98 \\
Reflect      & 93.44 & 88.25 & 94.54 & 87.98 & 90.44 & 90.71 \\
Consistency  & 90.98 & 91.26 & 93.17 & 90.98 & 93.00 & 89.89 \\
Ours         & 93.44 & 92.62 & 93.20 & 92.35 & 93.44 & 93.26 \\
\bottomrule
\end{tabular}
\caption{Performance comparison of DeepSeek-V3 and Qwen3-14B on MATH-500 across different numbers of prompting examples.}
\label{tab:deepseek_qwen_math_results}
\end{table*}
We conducted extended analysis on all in-context learning experiments in \S~\ref{section:in-context-learning}, specifically examining how performance changes as the number of demonstration examples increases, as illustrated in Table~\ref{tab:deepseek_qwen_math_results}. Based on the experimental results, the proposed method ("Ours") demonstrates strong and consistent performance across both DeepSeek-V3 and Qwen3-14B models, achieving the highest or competitive accuracy in all example-count settings. While increasing the number of in-context examples does not uniformly improve performance—and sometimes even degrades it, particularly for Qwen3-14B under methods like Polish and Reflect—our approach remains robust, showing no noticeable decline. DeepSeek-V3 generally outperforms Qwen3-14B in most scenarios, though Qwen3-14B benefits markedly from our method, especially with 15 examples where it reaches 93.26\% accuracy. These findings highlight the effectiveness and generalizability of our approach compared to existing prompting strategies.

\section{Prompts}
We provide the prompts of FACT-E in Table~\ref{tab:prompts_our}, the prompts of baselines are shown in ~\ref{tab:base_prompt} and \ref{tab:prompt_details}.

\section{LLM Usage Claim} 
\label{app:H}
In this paper, LLMs are utilized exclusively for the purpose of aiding and polishing writing. Their application is strictly confined to improving linguistic clarity, coherence, grammar, and style within
textual content. No additional functionalities are incorporated.

\begin{table*}[htbp]
\centering
\footnotesize
\label{tab:baseline_prompts}
\begin{tabular}{p{3cm}p{11.5cm}}
\toprule
\textbf{Method} & \textbf{Prompt Template} \\
\midrule
\textbf{zero shot} & 
Please answer the following question without any explanation. \\
& \\
& Please format your response as follows: \\
& \\
& Answer: Final numeric answer \\
& \\
& Question: \{question\} \\
& \\
& Answer: \\
\midrule
\textbf{In-context Learning} & 
Following the given examples and think step by step to solve the following question. \\
& First provide the reasoning process (CoT), then give the final numeric answer. \\
& Please format your response as follows: \\
& \\
& CoT: Step-by-step reasoning \\
& Answer: Final numeric answer \\
& \\
& Following the examples below: \\
& \\
& After reviewing the following examples, solve the new problem in the same way: \\
& \\
& \{Few\_shot\_examples\} \\
& \\
& Now solve the following question: \\
& \\
& Question: \{question\} \\
& \\
\midrule
\textbf{One Few-shot Example} &
Question: "Josh decides to try flipping a house. He buys a house for \$80,000 and then puts in \$50,000 in repairs. This increased the value of the house by 150\%. How much profit did he make?" \\
& \\
& CoT: \\
& Step 1: Calculate the original value of the house: \$80,000 \\
& Step 2: Calculate the increase in value due to repairs: 150\% of \$80,000 = 1.5 × 80,000 = \$120,000 \\
& Step 3: Calculate the total selling price of the house: \$80,000 + \$120,000 = \$200,000 \\
& Step 4: Calculate the total cost incurred by Josh: \$80,000 (purchase) + \$50,000 (repairs) = \$130,000 \\
& Step 5: Calculate the profit: \$200,000 (selling price) - \$130,000 (total cost) = \$70,000 \\
& \\
& Answer: 70000 \\
\bottomrule
\end{tabular}
\caption{Baseline prompts}
\label{tab:base_prompt}
\end{table*}

\begin{table*}[htbp]
\centering
\footnotesize
\label{tab:contrastive_prompt}
\begin{tabular}{p{2cm}p{12cm}}
\toprule
\textbf{Module} & \textbf{Prompt Template} \\
\midrule
\textbf{$\mathcal{C}_{\mathbf{S}}$} & 
\textbf{Question:} \{question\} \\
& \textbf{CoT:} \{cot\} \\
& \textbf{Answer:} \{ground\_truth\} \\
& \\
& \textbf{Task:} Determine whether the provided Chain of Thought (CoT) logically deduces the correct answer for the given question. Respond with \textbf{"True"} if the reasoning leads to the answer, or \textbf{"False"} if it does not. \\

\midrule
\textbf{The counterfactual chain generation} &
You are given a math question and its corresponding reasoning chain. \\
& This reasoning chain is divided into two parts: \\
& - The steps \textbf{before step t}, called `Chain before step t`. \\
& - The steps \textbf{after step t}, called `Chain after step t`. \\
& \\
& Your task is to generate a completely alternative reasoning chain after step t, directly reflecting the following error: \\
& \{error\} \\
& \\
& The alternative reasoning chain must: \\
& 1. Start exactly where the chain before step t ends, preserving earlier logic. \\
& 2. Modify the original continuation to reflect the specified error type. \\
& 3. Be logically coherent up to step t and introduce the assigned error naturally. \\
& 4. End with a final boxed answer, if the original did. \\
& \\
& \textbf{Input Format:} \\
& \\
& Question: \{question\} \\
& Chain before step t: \\
& \{before\_step\_flip\} \\
& Chain after step t: \\
& \{after\_step\_flip\} \\
& \\
& \textbf{Output Format:} \\
& \\
& Contrastive Chain After Step t: \\
\midrule
\textbf{$\mathcal{F}_\mathbf{S}$} & 
Choose the better option directly, without explaining your reasoning. \\
& \\
& \textbf{Question:} "{question}" \\
& \textbf{Previous reasoning (partial chain of thought):} \\
& \{before\_step\_flip\} \\
& \\
& Now evaluate which of the following two options is a more logical, coherent, and fluent continuation of the previous reasoning. \\
& The better option should follow naturally from the previous steps and maintain consistency in mathematical logic and style. \\
& \\
& \textbf{Option A:} \\
& \{before\_step\_flip\},\{after\_step\_flip\} \\
& \\
& \textbf{Option B:} \\
&  \{before\_step\_flip\},\{contrastive\_cot\_entry['cot']\} \\
& \\
& \textbf{Answer Choice:} [A/B/NA] \\
\bottomrule
\end{tabular}
\caption{Prompts used in our method.}
\label{tab:prompts_our}
\end{table*}

\begin{table*}[t!] 
\small
\centering
\renewcommand{\arraystretch}{1.5}
\begin{tabularx}{\textwidth}{l|c|X}
\toprule
\textbf{Method} & \textbf{Stage} & \textbf{Prompt Content and Some examples of demo} \\ \midrule
\textsc{Consistency} & --- & \textbf{Base Prompt:} Think step by step to solve the following question. First provide the reasoning process (CoT), then give the numeric final answer. \\
 & & \textbf{Format:} \par CoT: Step-by-step reasoning \par Answer: Final numeric answer \\
 & & \textbf{Example:} \{Example\} \\ \midrule
\textsc{Reflect} & 1 & (Same as the \textbf{Base Prompt} above) \\ \cmidrule{2-3}
 & 2 & \textbf{Reflection Prompt:} Based on the Chain-of-Thought (COT) reasoning and the answer you just provided, please reconsider the following question. Confirm the correctness of your prior answer, and then answer it again, also using the Chain-of-Thought (COT) format followed by the final answer. \\ \midrule
 \textsc{Denoise}& 1 &(Same as the \textbf{Base Prompt} above)  \\ \cmidrule{2-3}
& 2 & \textbf{Masking Process:} Replace specific tokens within the CoT reasoning with [MASK] to create a denoising objective. \par
 \textit{An Example:} Question: A curve is parameterized by $(x,y) = (t^3 + 7, -3t^2 - 6t - 5).$ Find the point the curve passes through at $t = 2.$ \par To find the point on [MASK] curve at [MASK] $ t = 2 $, [MASK] need [MASK] substitute $t = [MASK] [MASK]$ into the parameterization [MASK] for $x$ [MASK] $y$. [MASK] parameterization is given by: [MASK] $x = t^3 + 7$ $ y = -3t^2 - [MASK] - 5 $ First, we [MASK]$ x $ [MASK] when $ t $[MASK] [MASK] [MASK] [MASK] [MASK] [MASK] [MASK] + 7 = 8 + 7 = 15. Next, [MASK] calculate [MASK]$ y $ when $ [MASK] [MASK] 2 $ [MASK] $y = [MASK] - [MASK] - [MASK] = -3(4) - 12 - 5 = [MASK] - 12 [MASK] 5 [MASK] [MASK]$. Thus, [MASK] point on [MASK] curve at [MASK] $t = [MASK]$ is $( [MASK], y) = (15, -29)$. \\ \cmidrule{2-3}
& 3 & \textbf{Inference Template:} \par
 \textbf{Instruction:} Please reconstruct and improve the following reasoning, then solve the question. \par
 \textbf{Question:} \{question\} 
 \textbf{Reasoning:} \{masked\_cot\} \par
 \textbf{Task:} Complete the reasoning by filling in the masked parts ([MASK]), then provide the final answer. \par
 \textbf{Format:} \par
 CoT: Step-by-step reasoning \par
 Answer: Final numeric answer \\ \midrule
 \textsc{Polish} & 1 & (Same as the \textbf{Base Prompt} above) \\ \cmidrule{2-3}
 & 2 & \textbf{Polish Template:} \par
 \textbf{Context:} \par
 Question: \{question\} \par
 Original CoT: \{CoT\} \par
 Original Answer: \{answer\} \par
 \textbf{Instruction:} Based on your previous answer and CoT to this question, please rewrite new versions of the CoT to be more understandable and more relevant to the question. Don't omit any useful information, especially the numbers, and maintain their original meaning when polysemous words appear. \par
 \textbf{Format:} \par
 CoT: Step-by-step reasoning \par
 Answer: Final numeric answer \par
 \textbf{Example:}  \{Example\} \\ \bottomrule 
\end{tabularx}
\caption{The configuration of baseline methods (for MATH-500). \textsc{Consistency} samples the base prompt $N$ times to reach a consensus, while \textsc{Reflect} and \textsc{Polish} utilize a sequential two-stage process for self-correction and refinement, respectively. \textsc{Denoise} incorporates a token recovery task, where the masking procedure is a programmatic implementation-level operation rather than a textual prompt.}
\label{tab:prompt_details}
\end{table*}

\end{document}